\documentclass[11pt,a4paper,logo]{minnesotanlp}
\usepackage[numbers,compress]{natbib}

\usepackage[utf8]{inputenc} 
\usepackage[T1]{fontenc}    
\usepackage{hyperref}       
\usepackage{url}            
\usepackage{booktabs}       
\usepackage{amsfonts}       
\usepackage{nicefrac}       
\usepackage{microtype}      
\usepackage{xcolor}         

\usepackage{adjustbox}
\usepackage{colortbl}    
\usepackage{longtable}   
\usepackage{pifont}      
\usepackage[most]{tcolorbox}
\usepackage{xspace}
\usepackage{multirow}
\usepackage{enumitem}
\usepackage{capt-of}
\usepackage{listings}
\usepackage{listingsutf8}
\usepackage{wrapfig}
\usepackage{subcaption}
\usepackage{arydshln}

\usepackage{calligra}
\usepackage{aurical}      

\definecolor{lstbg}{HTML}{F7F7F7}
\definecolor{lstkw}{HTML}{0B5394}
\definecolor{lstcom}{HTML}{6A737D}
\definecolor{lststr}{HTML}{B07B00}
\lstdefinestyle{heldoutcode}{
    backgroundcolor=\color{lstbg},
    basicstyle=\ttfamily\scriptsize,
    keywordstyle=\color{lstkw}\bfseries,
    commentstyle=\color{lstcom}\itshape,
    stringstyle=\color{lststr},
    showstringspaces=false,
    breaklines=true,
    breakatwhitespace=false,
    columns=flexible,
    keepspaces=true,
    frame=single,
    framerule=0.3pt,
    rulecolor=\color{lstcom!40},
    xleftmargin=4pt,
    xrightmargin=4pt,
    aboveskip=4pt,
    belowskip=4pt,
    inputencoding=utf8,
    extendedchars=true,
    literate=%
      {²}{{$^{2}$}}1
      {¹}{{$^{1}$}}1
      {·}{{$\cdot$}}1
      {×}{{$\times$}}1
      {ő}{{\H{o}}}1
      {τ}{{$\tau$}}1
      {—}{{---}}1
      {•}{{\textbullet}}1
      {⁻}{{$^{-}$}}1
      {ⁿ}{{$^{n}$}}1
      {₀}{{$_{0}$}}1
      {₁}{{$_{1}$}}1
      {₂}{{$_{2}$}}1
      {₅}{{$_{5}$}}1
      {₋}{{$_{-}$}}1
      {ₙ}{{$_{n}$}}1
      {ℝ}{{$\mathbb{R}$}}1
      {→}{{$\rightarrow$}}1
      {∈}{{$\in$}}1
      {∞}{{$\infty$}}1
      {∫}{{$\int$}}1
      {≈}{{$\approx$}}1
      {≤}{{$\leq$}}1
      {≥}{{$\geq$}}1
      {★}{{$\star$}}1
      {不}{{[CJK]}}1
      {正}{{[CJK]}}1
      {解}{{[CJK]}}1,
}
\lstdefinestyle{heldoutsys}{
    style=heldoutcode,
    basicstyle=\ttfamily\scriptsize,
    breaklines=true,
}

\AtBeginDocument{\setlength{\LTcapwidth}{\textwidth}}

\definecolor{LightBlue}{RGB}{220,230,250}
\definecolor{LightRed}{RGB}{250,220,220}

\definecolor{Red}{rgb}{0.768, 0.054, 0.054}
\definecolor{Blue}{rgb}{0.152, 0.294, 0.925}
\definecolor{Green}{rgb}{0,0.4,0.7}

\usepackage{cleveref}
\crefname{section}{§}{§§}

\definecolor{prompt_colorbox}{HTML}{fd7e14}
\newtcolorbox{prompt}[2][]{
    colback=white,
    colframe=prompt_colorbox!45,
    fonttitle=\bfseries,
    coltitle=black,
    sharp corners,
    title=#2,
    #1,breakable
}

\usepackage{color-edits}
\addauthor{gn}{magenta}

\newcommand{\eg}{e.g.,\xspace}
\newcommand{\ie}{i.e.,\xspace}

\newcommand{\cmark}{\textcolor{green!70!black}{\ding{51}}} 
\newcommand{\xmark}{\textcolor{red}{\ding{55}}} 

\newcommand{\diff}{\texttt{diff-based edit}\xspace}
\newcommand{\full}{\texttt{full rewrite}\xspace}

\newcommand{\pythonicon}{\raisebox{-0.2ex}{\includegraphics[height=1em]{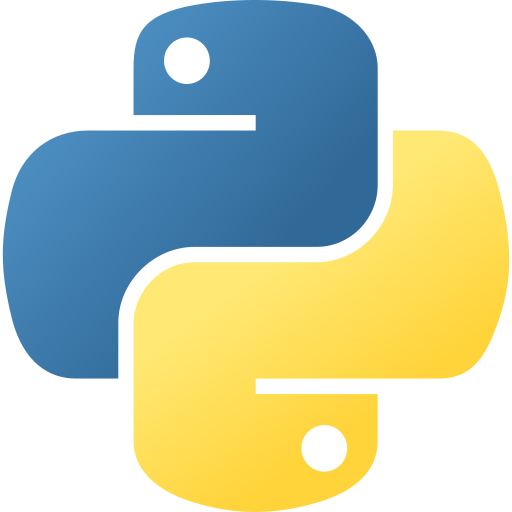}}}
\newcommand{\cppicon}{\raisebox{-0.2ex}{\includegraphics[height=1em]{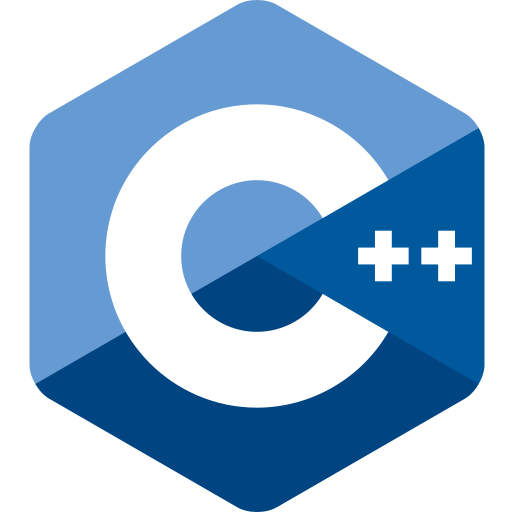}}}
\newcommand{\finchicon}{\raisebox{-0.2ex}{\includegraphics[height=1em]{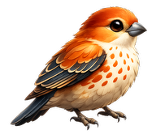}}}

\DeclareMathOperator*{\argopt}{arg\,opt}

\definecolor{finchcolor}{HTML}{EE8C3C}
\newcommand{\finchcolor}{\rowcolor{finchcolor!20}}

\newcommand{\good}[1]{\textcolor{teal}{#1}}
\newcommand{\bad}[1]{\textcolor{red!70!black}{#1}}
\newcommand{\deltarowcolor}{\rowcolor{gray!10}}

\definecolor{finchOrange}{HTML}{D94E1F}    
\definecolor{finchCharcoal}{HTML}{3D3935}  
\definecolor{finchCream}{HTML}{FAEFD9}     
\definecolor{finchAccent}{HTML}{E8843D}    

\definecolor{improvedcolor}{HTML}{C97A4A}   
\definecolor{nochangecolor}{HTML}{E8B98A}   
\definecolor{regressedcolor}{HTML}{3A2F2A}  

\usepackage{xspace}

\newcommand{\datasetName}{\ensuremath{\mathcal{F}}inch Collection\xspace}
\newcommand{\modelName}{\ensuremath{\mathcal{F}}inch\xspace}

\newcommand{\linedlabel}[2]{%
  {\color{#1}%
    \setbox0=\hbox{\texttt{#2}}%
    \rule[\dimexpr\ht0+1pt]{\wd0}{0.5pt}%
    \kern-\wd0%
    \texttt{#2}%
    \kern-\wd0%
    \rule[-2pt]{\wd0}{0.5pt}%
  }%
}

\newcommand{\improved}{\linedlabel{improvedcolor}{\texttt{Imp}}\xspace}
\newcommand{\nochange}{\linedlabel{nochangecolor!80!black}{\texttt{NC}}\xspace}
\newcommand{\regressed}{\linedlabel{regressedcolor}{\texttt{Reg}}\xspace}

\title{\textsc{Evolution Fine-Tuning}: Learning to Discover Across 371 Optimization Tasks}

\author[1]{Young-Jun Lee}
\author[2]{Seungone Kim}
\author[3]{Minki Kang}
\author[2]{Alistair Cheong}
\author[4]{Zerui Chen}
\author[5]{Seungho Han}
\author[6]{Taehee Jung\thanks{This work is independent of the author's position at Amazon and does not relate to any work conducted at Amazon.}}
\author[1]{Dongyeop Kang}
\affil[1]{University of Minnesota}
\affil[2]{Carnegie Mellon University}
\affil[3]{KAIST}
\affil[4]{University of Cambridge}
\affil[5]{Hanyang University}
\affil[6]{Amazon}

\newcommand{\projectbadges}{%
  \raisebox{-0.4ex}{\includegraphics[height=1em]{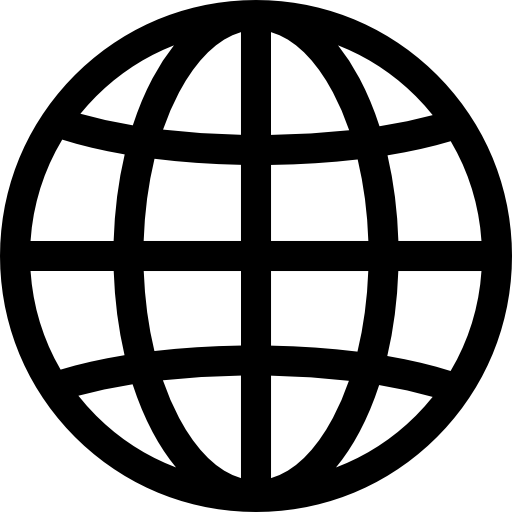}}\hspace{0.3em}\href{https://open-galapagos.github.io/evolution_finetuning/}{\texttt{Website}}\hspace{0.6cm}%
  \raisebox{-0.4ex}{\includegraphics[height=1em]{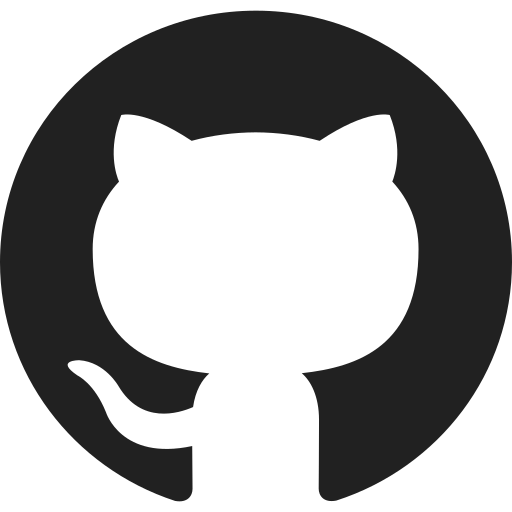}}\hspace{0.3em}\href{https://github.com/Open-Galapagos/evolution-fine-tuning}{\texttt{Code}}\hspace{0.6cm}%
  \raisebox{-0.4ex}{\includegraphics[height=1em]{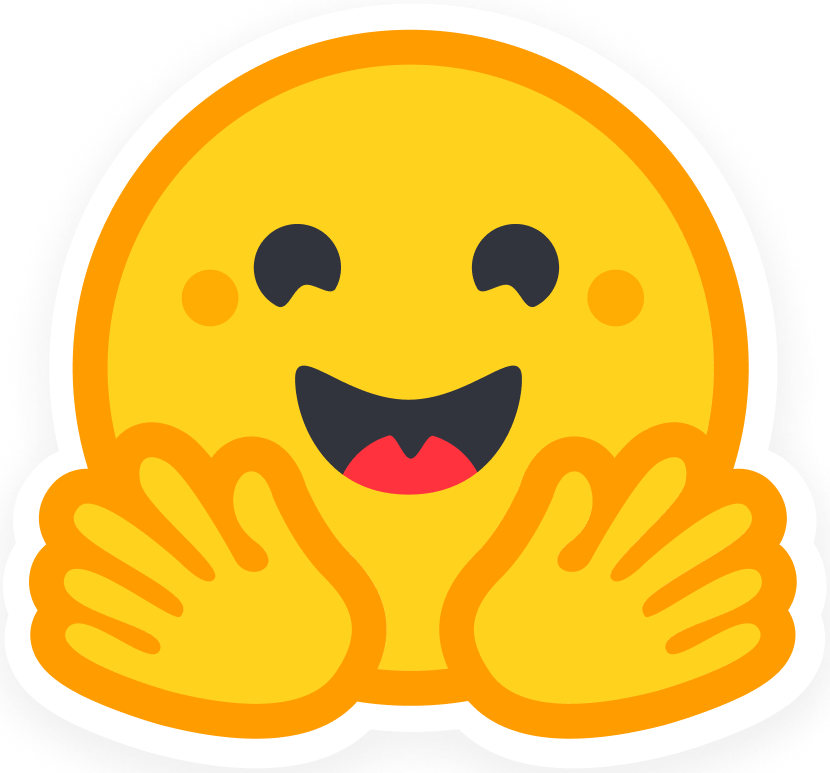}}\hspace{0.3em}\href{https://huggingface.co/collections/minnesotanlp/evolution-fine-tuning}{\texttt{Dataset \& Models}}%
}
\renewcommand{\titleextra}{\projectbadges}

\begin{document}

\begin{abstract}
    Would experience designing faster GPU kernels also help close in on a long-standing open mathematical conjecture?
    Large Language Models (LLMs) integrated into evolutionary search have recently produced state-of-the-art solutions on optimization tasks, including open mathematical conjectures, GPU kernel design, scientific law discovery, and combinatorial puzzles.
    To achieve this, prior work applied search scaffolds to one target task at a time, so every new problem is approached from scratch and the experience accumulated during search is discarded once the model finishes its attempt. 
    This leaves the capability of iteratively evolving a solution (\textit{e.g.}, knowing which part to mutate and how, deciding when to backtrack) entirely in the scaffold rather than in the model itself. 
    Whether the model itself could acquire this capability and reuse it across different tasks has been largely unexamined.
    To address this, we introduce \textbf{\textsc{Evolution Fine-Tuning} (EFT)}, a mid-training paradigm that teaches LLMs to evolve solutions across tasks by converting evolutionary search trajectories into supervision.
    We construct \datasetName, a 156K-trajectory dataset spanning 10 domains and 371 optimization tasks, and fine-tune open-source LLMs from 2B to 9B parameters.
    Empirically, EFT confers cross-task generalization: across 22 held-out tasks, our models surpass their base counterparts by 10.22\% on average. 
    Furthermore, when paired with test-time RL, our model matches state-of-the-art performance on two circle-packing tasks and outperforms its base-model counterpart on the Erdős minimum-overlap problem. EFT thus serves as a ``practice phase'' for general-purpose discovery agents that do not solve new problems from scratch.
\end{abstract}

\maketitle
\vspace{-3mm} 

\begin{figure}[!h]
    \centering
    \includegraphics[width=\linewidth]{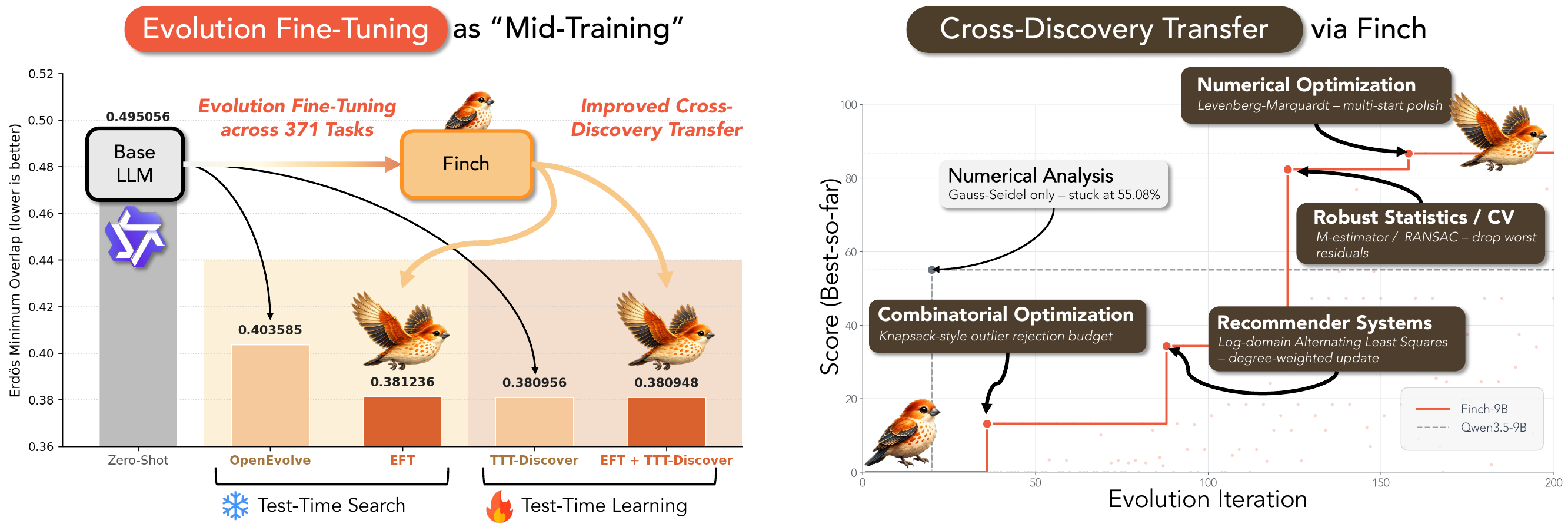}
    \caption{\textbf{(Left)} \textbf{\textsc{Evolution Fine-Tuning} (EFT)} serves as ``mid-training'', boosting \modelName's discovery capability on the Erd\H{o}s minimum overlap problem under both test-time search and learning. \textbf{(Right)} On NP-hard competitive programming (CALICO, UC Berkeley contest), EFT enables cross-discovery transfer: \modelName solves problems by combining strategies acquired from diverse domains, such as combinatorial optimization, recommender systems, robust statistics/computer vision, and numerical optimization. In contrast, the base model without EFT relies on a single, repetitive strategy.}
    \label{main_fig:intro_teaser}
\end{figure}

\section{Introduction} \label{main_sec:introduction}

Some of the most consequential problems in mathematics, algorithm engineering, and the natural sciences are optimization tasks: problems for which a candidate solution can be scored against an objective, but for which the optimal solution is not directly computable~\citep{kirkpatrick1983optimization}. 
Concrete examples include open mathematical conjectures such as the Erdős minimum-overlap problem~\citep{erdHos1955some, white2023new}, the design of high-performance GPU kernels~\citep{ouyang2025kernelbench}, and the discovery of new scientific laws from data~\citep{shojaee2025llm}.

Recently, large language models (LLMs) combined with evolutionary search methods have begun to produce state-of-the-art solutions across such tasks: at each iteration, the LLM proposes new candidate solutions, a scaffold scores them and updates a population of high-scoring candidates, and the loop continues until a strong solution emerges~\citep{romera2024mathematical,novikov2025alphaevolve,lange2025shinkaevolve}.
Two methodological branches dominate this line of work: 
(1) \emph{Test-time search} methods use a fixed, typically proprietary LLM as the mutation operator and rely on the scaffold's parent selection and prompting logic to drive improvement~\citep{assumpccao2025codeevolve,yan2026pacevolve,cemri2026adaevolve,liu2026evox}.
(2) \emph{Test-time learning} methods additionally update the LLM's weights during the search process, allowing the model to specialize to the target task as it explores~\citep{wang2025thetaevolve,yuksekgonul2026learning}.

Despite empirical successes, both branches share a common limitation: the discovery capability (\textit{i.e.}, the skill of iteratively improving a solution, knowing what to mutate, what to keep, and when to backtrack) is constructed during each search rather than internalized into the model itself. Specifically,
\begin{enumerate}[leftmargin=*]
    \item Test-time search methods often rely on proprietary, frontier-scale LLMs as their mutation operator, because the scaffold demands consistent, high-quality proposals at every iteration. In our experiments, we observe that open-source models smaller than 9B parameters fail to follow evolutionary trajectories within such scaffolds and yield substantially weaker performance (Figure~\ref{main_fig:intro_teaser}, left).
    \item Test-time learning methods alleviate this by allowing smaller LLMs to adapt their weights based on their own search experience, and have produced new best-known solutions on several mathematical problems~\citep{wang2025thetaevolve, yuksekgonul2026learning}. However, these updates are tailored to a single search loop and a single task; the strategies the model discovers are not consolidated into reusable capability, so the model cannot compose strategies from prior tasks when tackling a new one (Figure~\ref{main_fig:intro_teaser}, right).
    \item More fundamentally, in neither branch does the model itself acquire the evolving capability. Test-time search-based methods do not update the model at all, leaving the capability in the search procedure by design. Test-time learning-based methods update the model through test-time RL, but the updates serve to find a solution within a single search loop rather than to internalize the discovery capability itself, and they are discarded once the task is solved.
\end{enumerate}

One promising direction to address these limitations is to have the LLM itself \textit{meta-learn} the discovery capability (\textit{i.e.}, learning how to evolve solutions across optimization tasks). 
The core challenge in doing so is that optimization tasks are NP-hard and lack ground truth optimal solutions, making the standard supervised learning recipe of collecting (problem, answer) pairs unavailable. 
To circumvent this, we propose \textbf{\textsc{Evolution Fine-Tuning (EFT)}}, a mid-training paradigm that treats the \textit{trajectories} of search runs as the supervision signal, thereby internalizing the discovery capability into the model itself. 
We construct \datasetName, a large-scale dataset of 156K such trajectories collected using a widely used search scaffold, \ie OpenEvolve~\citep{openevolve}, and a recent large model, Qwen3.5-397B-A17B~\citep{qwen35blog}, across 10 domains and 371 tasks.
Using \datasetName, we fine-tune open-source models from 2B to 9B, and obtain a new model family, \finchicon \modelName-\{2, 4, 8, 9\}B.

\begin{wrapfigure}{r}{0.6\linewidth}
    \centering
    \vspace{-5mm}
    \includegraphics[width=\linewidth]{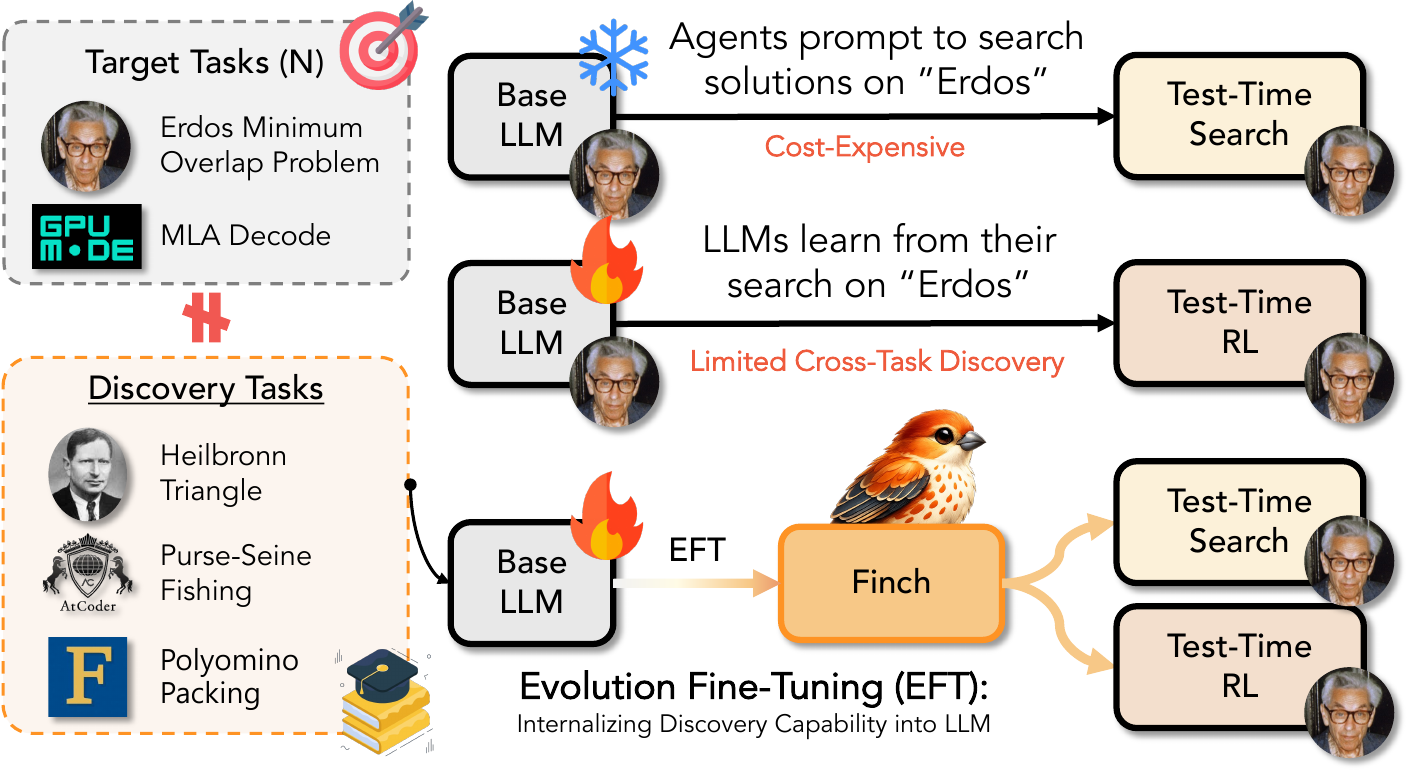}
    \vspace{-6mm}
    \captionsetup{justification=centering}
    \caption{A Concept of Evolution Fine-Tuning (EFT)}
    \vspace{-3mm}
    \label{main_fig:eft_concept}
\end{wrapfigure}

To demonstrate the cross-task generalization conferred by \datasetName, we first employ \modelName as mutation operators in test-time search scaffolds. \modelName outperforms its base counterparts on 22 held-out tasks and achieves performance comparable to best-known solutions previously obtained by a proprietary model, despite using a much smaller open-source backbone. Notably, as shown in Figure~\ref{main_fig:intro_teaser} (right), when solving a competitive programming task, we observe that while base LLM tends to apply only in-domain strategy (\ie Gauss-Seidel uniform weight), \modelName transfers strategies across domains (\eg applying log-domain alternating least squares from recommender system, Levenberg-Marquardt from numerical optimization to solve a competitive programming problem), suggesting that \textbf{EFT} gives rise to emergent behaviors in discovery tasks. Furthermore, scaling the number of training tasks in \datasetName from 15 to 355 improves \modelName's held-out performance by \textbf{14.1\%} on average across held-out tasks. Finally, to assess whether \modelName can also learn from its own search experience, we apply test-time learning to both \modelName (\ie w/ \textbf{EFT}) and base LLMs (\ie w/o \textbf{EFT}) on three tasks. We find that \modelName achieves state-of-the-art performance on two circle packing tasks and outperforms its base-model counterpart on the Erdős Minimum Overlap Problem.

\section{Preliminaries} \label{main_sec:preliminaries}

\paragraph{Optimization setup.}
We consider an optimization task $\tau \in \mathcal{T}$ with an initial candidate solution $x_0$ and an iteration budget $T$.
The candidate set generated during search is denoted by $\mathcal{X}=\{x_0,\ldots,x_T\}$, where $x_0$ may be a program \citep{novikov2025alphaevolve}, math construction \citep{imajuku2025ale}, or prompt \citep{openevolve}, depending on the task.
At iteration $t$, an evolutionary scaffold $\mathcal{S}$ uses a mutation operator $\mathcal{M}_{\theta}$, typically an LLM, to produce a new candidate from the parent solution and search history:
    $x_t = \mathcal{S}(x_{t-1}, I, \mathcal{H}_{t-1}; \mathcal{M}_{\theta})$,
where $\mathcal{H}_{t-1}$ contains selected prior candidates and feedback. An evaluator $\mathcal{E}$ assigns a score and auxiliary artifacts such as logs or natural-language feedback. The goal is
\begin{equation}
    x^{\star}=\argopt_{x\in\mathcal{X}}\mathcal{E}(x),
    \qquad \mathrm{opt}\in\{\max,\min\},
    \label{eq:objective}
\end{equation}
where the optimization direction is determined by task: $\max$ for accuracy and $\min$ for c5 bound in Erdos problem.

\begin{figure}[t!]
\centering
\begin{minipage}{0.62\linewidth}
\centering
\captionof{table}{Comparison of evolutionary methods. 
Scaffold indicates whether the method provides search or learning support for evolution. 
Train and Test denote scaffolding at training or test time, respectively. 
OS denotes open sourcing.}
\label{main_tab:method_comparison}
\vspace{3pt}
\setlength{\tabcolsep}{2.5pt}
\renewcommand{\arraystretch}{1.08}
\footnotesize
\begin{adjustbox}{max width=\linewidth}
\begin{tabular}{@{}l c cc cc c c c@{}}
\toprule
\multirow{2}{*}{\textbf{Method}} &
\multirow{2}{*}{\shortstack{\textbf{Main}\\\textbf{Contribution}}} &
\multicolumn{2}{c}{\textbf{Scaffold}} &
\multicolumn{4}{c}{\textbf{Training}} &
\multirow{2}{*}{\textbf{OS}} \\[-3pt]
\cmidrule(lr){3-4}
\cmidrule(lr){5-8}
&
&
{Search} &
{Learn} &
{Train} &
{Test} &
{Paradigm} &
{\#Tasks} &
\vspace{-1mm}
\\
\midrule
AlphaEvolve~\citep{novikov2025alphaevolve}
& Search
& \cmark & \xmark & \xmark & \xmark & -- & -- & \xmark \\
OpenEvolve~\citep{openevolve}
& Search
& \cmark & \xmark & \xmark & \xmark & -- & -- & \cmark \\
ShinkaEvolve~\citep{lange2025shinkaevolve}
& Search
& \cmark & \xmark & \xmark & \xmark & -- & -- & \cmark \\
GEPA~\citep{agrawal2025gepa}
& Search
& \cmark & \xmark & \xmark & \xmark & -- & -- & \cmark \\
PAC-Evolve~\citep{yan2026pacevolve}
& Search
& \cmark & \xmark & \xmark & \xmark & -- & -- & \xmark \\
AdaEvolve~\citep{cemri2026adaevolve}
& Search
& \cmark & \xmark & \xmark & \xmark & -- & -- & \cmark \\
EvoX~\citep{liu2026evox}
& Search
& \cmark & \xmark & \xmark & \xmark & -- & -- & \cmark \\
DGM~\citep{zhang2025darwin}
& Search
& \cmark & \xmark & \xmark & \xmark & -- & -- & \cmark \\
HyperAgent~\citep{zhang2026hyperagents}
& Search
& \cmark & \xmark & \xmark & \xmark & -- & -- & \cmark \\
CORAL~\citep{qu2026coral}
& Search
& \cmark & \xmark & \xmark & \xmark & -- & -- & \cmark \\
\hline
ThetaEvolve~\citep{wang2025thetaevolve}
& Learning
& \xmark & \cmark & \xmark & \cmark & RL & 1 & \cmark \\
TTT-Discover~\citep{yuksekgonul2026learning}
& Learning
& \xmark & \cmark & \xmark & \cmark & RL & 1 & \cmark \\
\midrule
\rowcolor{gray!10}
\textbf{EFT (Ours)}
& \textbf{Model, data}
& \cmark & \cmark & \cmark & \cmark & \textbf{SFT, RL} & \textbf{371} & \cmark \\
\bottomrule
\end{tabular}
\end{adjustbox}
\end{minipage}%
\hfill
\begin{minipage}{0.36\linewidth}
    \centering
    \includegraphics[width=\linewidth]{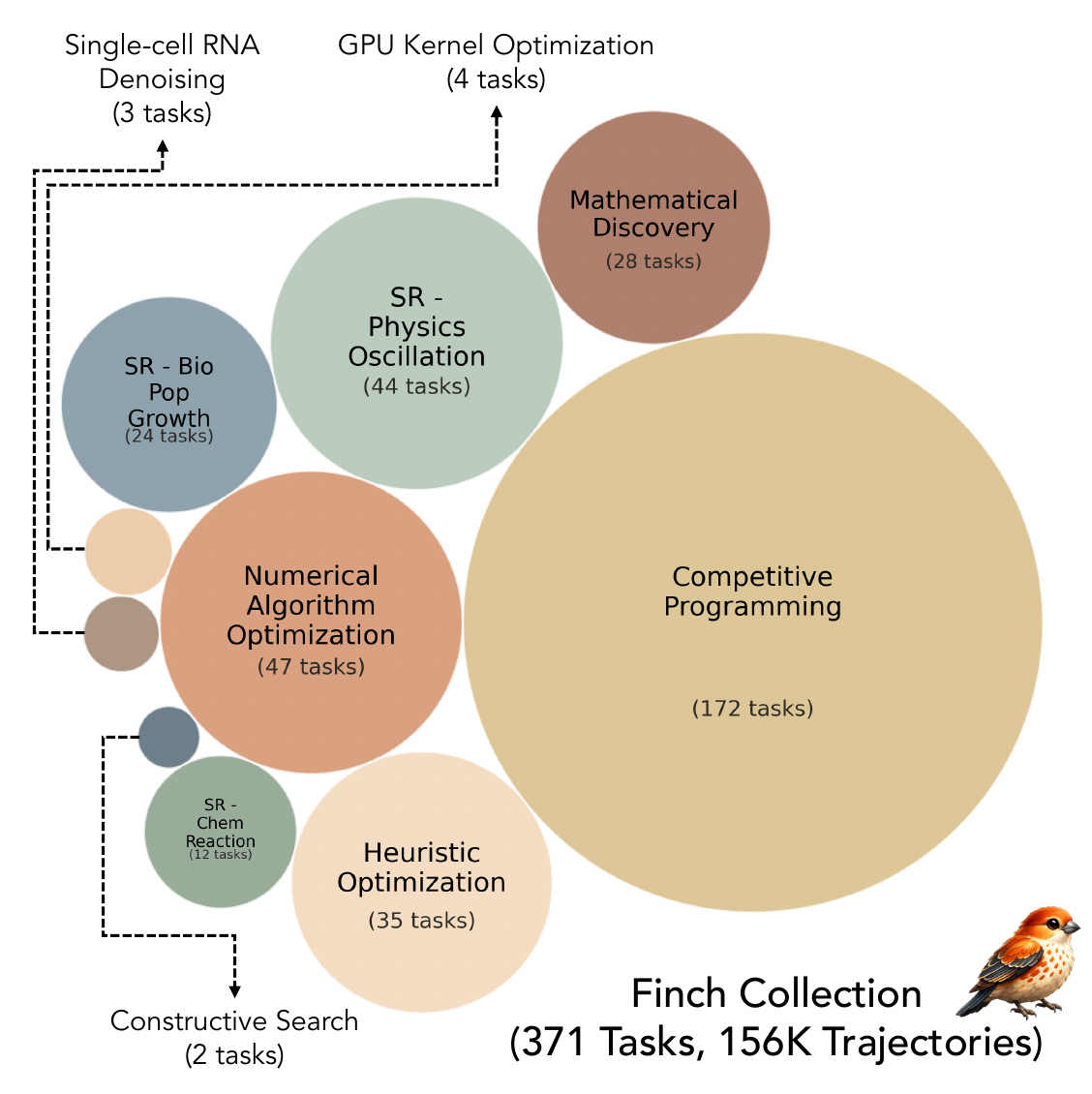}
    \captionof{figure}{An overview of the optimization task groups (a total of 371 tasks) in \datasetName, where bubble size indicates the number of tasks in each group.}
    \label{main_fig:task_group_dist}
\end{minipage}
\end{figure}

\paragraph{Discovery.}
Following prior work~\citep{yuksekgonul2026learning}, we call $x^{\star}$ a \textit{discovery} if it improves upon the previous best-known solution $x_{\mathrm{sota}}$ within budget $T$, i.e.,
$\mathcal{E}(x^{\star})>\mathcal{E}(x_{\mathrm{sota}})$ for maximization tasks, with the inequality reversed for minimization tasks.

\paragraph{Evolutionary Search Scaffold.}
To discover novel solutions $\tau$, existing LLM-based evolutionary methods use either search-based or learning-based scaffolds. Search-based scaffolds keep $\theta$ fixed and rely on external search mechanisms, while learning-based scaffolds update $\theta$ at test time. Both approaches are fundamentally built upon four core modular components: (i) construct a prompt from the parent solution, task instruction, history, and feedback; (ii) generate a candidate using $\mathcal{M}_{\theta}$ through either \texttt{diff-based edit} or \texttt{full rewrite}; (iii) evaluate the candidate with $\mathcal{E}$; and (iv) store eligible candidates in a population database $\mathcal{D}$. The updated database $\mathcal{D}_t$ then informs the next iteration until the compute budget or target score is reached.
We summarize existing evolutionary methods in Tab \ref{main_tab:method_comparison}.

\section{\textsc{Evolution Fine-Tuning}}

Evolutionary scaffolds can discover strong solutions at test time, but the discovery procedure is typically external to the model. We introduce \textbf{Evolution Fine-Tuning (EFT)}, a mid-training procedure that transfers this test-time discovery behavior into smaller open-source LLMs. EFT converts evolutionary search trajectories into supervised training examples, so that the model learns to act as a stronger mutation operator before deployment. As summarized in Table~\ref{main_tab:method_comparison}, EFT is orthogonal to the choice of test-time scaffold: the resulting model can be used inside search-based scaffolds with frozen weights, or further adapted by learning-based scaffolds such as test-time RL.

\begin{figure}[!t]
    \centering
    \includegraphics[width=\linewidth]{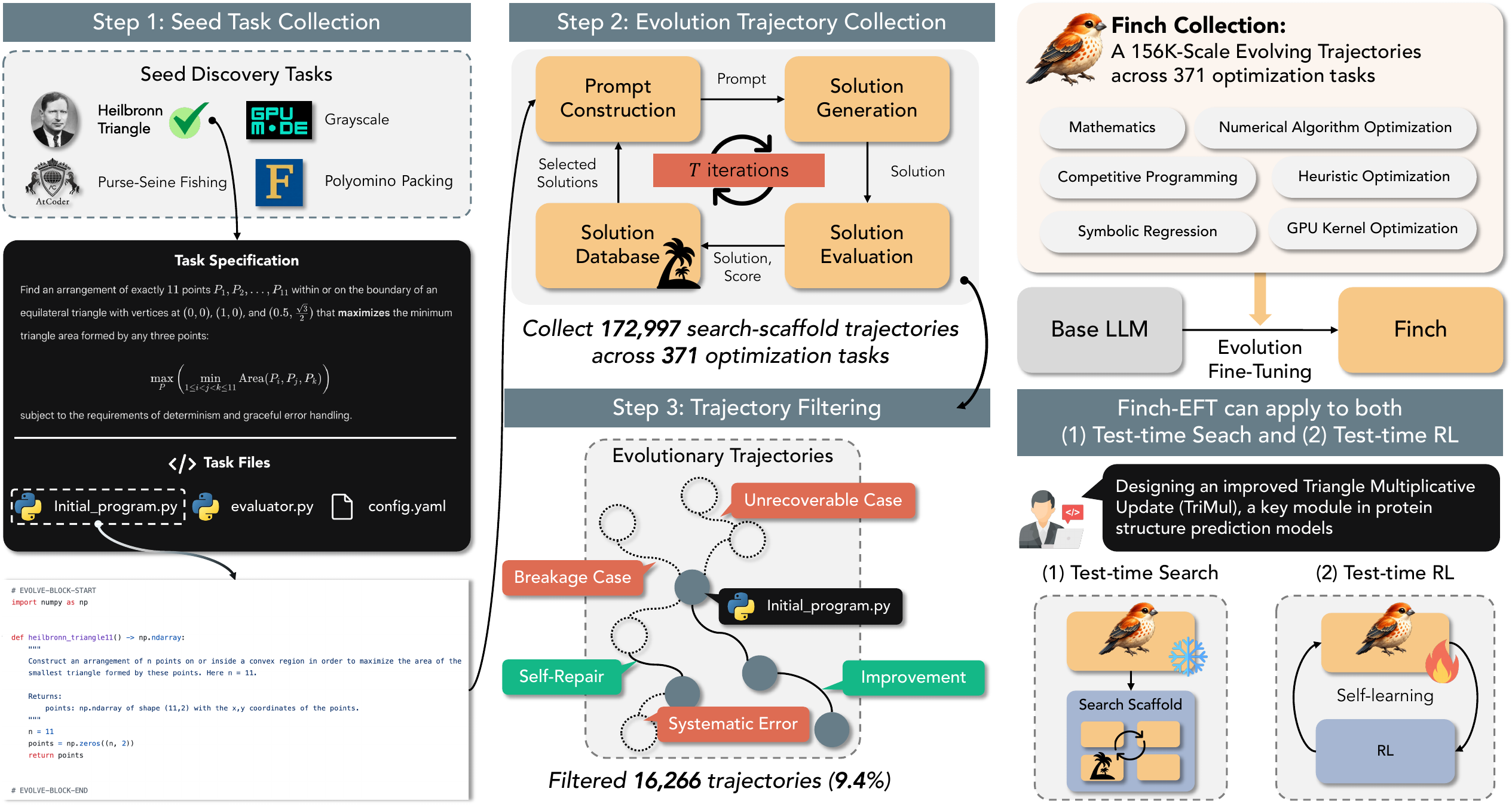}
    \caption{Overview of the \datasetName construction pipeline, consisting of (1) seed optimization task collection, (2) trajectory collection via an evolutionary search scaffold (\ie OpenEvolve~\citep{openevolve}), and (3) trajectory filtering for unrecoverable cases, breakage cases, and candidate solutions that incur systematic errors (\eg timeout errors).}
    \label{main_fig:method_overview}
\end{figure}

\subsection{\datasetName Construction} \label{main_sec:dataset_construction}

\paragraph{Overview.}
Figure~\ref{main_fig:method_overview} illustrates the construction of \datasetName. Each task consists of a task specification, an initial candidate solution, an evaluator, and configuration files. We run an evolutionary scaffold over these task files to produce a sequence of parent-to-child solution transitions. Each transition records the prompt, parent solution, generated candidate, evaluator output, score change, and execution artifacts. We then remove trajectories whose feedback is unreliable or whose transitions would provide misleading supervision. The final collection contains approximately 156K evolutionary trajectories across 371 optimization tasks.

\paragraph{Step 1: Seed Optimization Task Collection.} 
Training data for optimization is difficult to synthesize because many target problems are NL-hard, lack known global optima, and require expert-designed evaluators (see Table~\ref{supp_tab:task_difference} in Appendix~\ref{supp_sec:discussion}). 
Instead of generating artificial tasks, we source seed tasks from existing optimization benchmarks whose objectives are executable and externally validated.
We select tasks according to three criteria: (i) the task should require nontrivial search, (ii) should not reduce to matching a known ground-truth answer, and (iii) should provide a deterministic evaluator that assigns a continuous or comparable score to candidate solutions.

In total, as shown in Figure~\ref{main_fig:task_group_dist}, we collect 371 seed tasks from 10 benchmarks, spanning from mathematical discovery, competitive programming, heuristic optimization, numerical algorithm optimization, symbolic regression, GPU kernel optimization, constructive search, and biological denoising benchmarks. 
These include AlphaEvolve's mathematical discovery problems~\citep{novikov2025alphaevolve}, FrontierCS~\citep{mang2025frontiercs}, ALE-Bench~\citep{imajuku2025ale}, AlgoTune~\citep{press2025algotune}, GPU Mode~\citep{gpumode}, LLM-SRBench~\citep{shojaee2025llm}, Function Minimization and K-Module tasks from OpenEvolve~\citep{openevolve}, scRNA-seq denoising~\citep{luecken2025defining,yuksekgonul2026learning}, and variants of Erd\H{o}s problems~\citep{feng2026semi,erdosproblems}. The full task list is provided in Appendix~\ref{supp_sec:full_task_list}.

\paragraph{Step 2: Evolutionary Trajectory Collection.}
For each seed task $\tau$, we run an evolutionary scaffold $\mathcal{S}$ for a fixed budget of iterations. At iteration $t$, the scaffold selects a parent solution $x_{t-1}$ and constructs a prompt from the task instruction $I$, the parent, the search history $\mathcal{H}_{t-1}$, and evaluator artifacts. 
For each task, $N_\tau$ trajectories are generated where $N_\tau < T$, where trajectories flagged as errors are discarded. A teacher mutation operator $\mathcal{M}_{\theta}$ then generates a candidate solution $x_t$, which is executed by the task evaluator $\mathcal{E}$. The resulting trajectory stores
    $(I, x_{t-1}, \mathcal{H}_{t-1}, x_t, \mathcal{E}(x_t), \mathcal{F}_t)$,
where $\mathcal{F}_t$ includes execution logs, error traces, and evaluator feedback.

We instantiate $\mathcal{M}$ with OpenEvolve~\citep{openevolve} and use Qwen3.5-397B-A17B~\citep{qwen35blog} as the teacher mutation operator.
To expose the student model to both local refinement and global exploration, we  collect trajectories under two mutation strategies: \diff and \full. The \diff strategy asks the model to edit an existing solution and therefore emphasizes exploitation, while \full asks the model to rewrite the solution and therefore encourages broader exploration. For diversity, we run the scaffold multiple times per task with stochastic decoding. Unless otherwise specified, we use temperature $0.7$, top-$p=0.95$, and a maximum generation length of 30K tokens. 
In total, we obtain 172,997 trajectories.

\paragraph{Step 3: Trajectory Filtering.}
Raw evolutionary traces contain failures that should not be imitated.
From 172,997 raw trajectories, we retain 156,731 (90.6\%) by applying the following criteria:
\begin{itemize}[leftmargin=2em, itemsep=1pt, parsep=0pt, topsep=1pt]
    \item First, we remove \textbf{systematic errors}, where the evaluator output is unreliable or the score delta cannot be computed, removing 6,321 trajectories (3.7\%) in total.
    Examples include missing parent scores (57.5\%), timeout errors (14.2\%), syntax-check failures, import failures, and evaluator crashes. 
    The complete list of systematic errors is provided in Table~\ref{supp_tab:system_error_breakdown}.
    \item Second, we remove \textbf{unrecoverable \& breakage cases}: Unrecoverable cases are both parent and child solutions are erroneous (294; 0.2\%), and breakage cases are an error-free parent yields an erroneous child (1,281; 0.8\%). Both constitute hard negatives, destabilizing the training signal.
    \item Lastly, we remove \textbf{excessively long inputs}, to keep training stable and tractable.
    We dsicard responses longer than 16,384 tokens and examples whose total serialized input-output length exceeds 32,768 tokens, removing 8,370 (5.0\%).
\end{itemize}
After filtering, \datasetName contains approximately 156K trajectories across 371 tasks.

\begin{wrapfigure}{r}{0.35\linewidth}
    \centering
    \vspace{-2em}
    \includegraphics[width=0.8\linewidth]{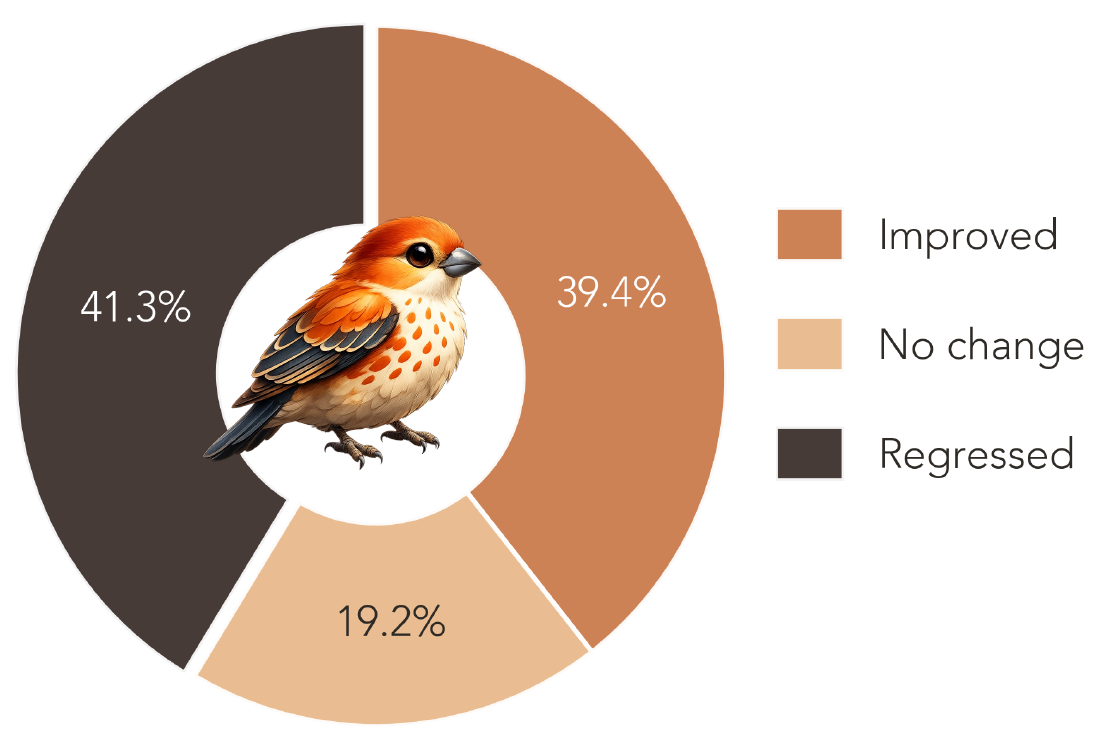}
    \caption{Distribution of trajectory improvement in \datasetName.}    
    \label{main_fig:finch_collection_improvement_dist}
    
    \setlength{\tabcolsep}{4pt}
    \renewcommand{\arraystretch}{1.3}
    \captionof{table}{Effect of improvement type on EFT.}
    \vspace{2pt}
    \label{main_tab:effect_of_improvement}
    \begin{adjustbox}{width=\linewidth}
    \begin{tabular}{lccc}
    \toprule
    \textbf{Model} & \textbf{Erdos} ($\downarrow$) & \textbf{AC1} ($\downarrow$) & \textbf{AC2} ($\uparrow$) \\
    \midrule
    Qwen3-8B & 0.403585 & 1.5177 & 0.8980 \\
    \hdashline
    + \improved & \textbf{0.381236} & \textbf{1.5154} & \textbf{0.9001} \\
    + \improved + \nochange + \regressed & 0.492126 & 1.5186 & 0.8789 \\
    \bottomrule
    \end{tabular}
    \end{adjustbox}
    \vspace{-1em}
\end{wrapfigure}

Each retained trajectory is converted into a supervised fine-tuning instance.
The input consists of the same information available to the mutation operator inside the scaffold: the task instruction, parent solution, selected history, previous scores, and evaluator artifacts. The target output is the teacher-generated candidate solution. This format directly trains the model to map an evolutionary state to a plausible next mutation: $(I, x_{t-1}, \mathcal{H}_{t-1}, \mathcal{F}_{t-1}) \mapsto x_t$.

When task scores are available, we classify each trajectory based on its improvement outcome ($\Delta$), \ie $\Delta = \mathcal{E}(x_t) - \mathcal{E}(x_{t-1})$, into three categories: $\Delta > 0$ (\improved), $\Delta = 0$ (\nochange), and $\Delta < 0$ (\regressed). As shown in Figure~\ref{main_fig:finch_collection_improvement_dist}, this yields 61,802 (39.4\%) \improved, 30,130 (19.2\%) \nochange, and 64,799 (41.3\%) \regressed trajectories. In this work, we primarily use \improved trajectories for evolution fine-tuning to prevent \modelName from imitating non-improving behaviors due to \regressed, as shown in Table~\ref{main_tab:effect_of_improvement}. Furthermore, to enable \modelName to learn how to evolve parent programs effectively within the solution space, we leverage both \improved and \regressed trajectories through a preference learning algorithm (\ie KTO~\citep{ethayarajh2024kto})~\footnote{To maximize the contrastive signal that guides \modelName toward self-judging which solutions are promising and which fall short, we restrict KTO training to \improved and \regressed in this work. Nevertheless, we believe \nochange is also a valuable resource for internalizing discovery capability, as suggested in Table~\ref{main_tab:hardsoft1}.}. Consequently, our \datasetName provides a comprehensive set of improvement trajectories, serving as a valuable resource for internalizing discovery capabilities into LLMs.

\begin{figure}[t!]
\centering
\setlength{\tabcolsep}{4pt}
\renewcommand{\arraystretch}{0.95}

\begin{minipage}[t]{0.50\linewidth}
    \vspace{0pt}
    \centering

    \resizebox{0.9\linewidth}{!}{%
    \begin{tabular}{@{}lcc@{}}
        \toprule
        \textbf{Task Group} & \textbf{\# Tasks} & \textbf{\# Traj.} \\
        \midrule
        Competitive Programming          & 172 & 43{,}469 \\
        Numerical Algorithm Optimization & 47  & 4{,}835  \\
        SR Physics Oscillation           & 44  & 51{,}210 \\
        Heuristic Optimization           & 35  & 5{,}924  \\
        Mathematical Discovery           & 28  & 6{,}508  \\
        SR Bio Pop Growth                & 24  & 28{,}034 \\
        SR Chem Reaction                 & 12  & 13{,}975 \\
        GPU Kernel Optimization          & 4   & 1{,}088  \\
        Single cell RNA Denoising        & 3   & 503      \\
        Constructive Search              & 2   & 1{,}185  \\
        \midrule
        \textbf{Total}                   & \textbf{371} & \textbf{156{,}731} \\
        \bottomrule
    \end{tabular}}

    \vspace{0.3em}
    {\footnotesize (a) Tasks and trajectories by task group}

    \vspace{0.7em}

    \resizebox{0.92\linewidth}{!}{%
    \begin{tabular}{lc|lc}
        \toprule
        \textbf{Language} & \textbf{Ratio} & \textbf{Mutation} & \textbf{Ratio} \\
        \midrule
        \pythonicon~Python & 68.5\% & \diff & 50.3\% \\
        \cppicon~C++       & 31.5\% & \full & 49.7\% \\
        \bottomrule
    \end{tabular}}

    \vspace{0.3em}
    {\footnotesize (b) Languages and mutation strategies}
\end{minipage}
\begin{minipage}[t]{0.48\linewidth}
    \vspace{0pt}
    \centering

    \includegraphics[width=\linewidth]{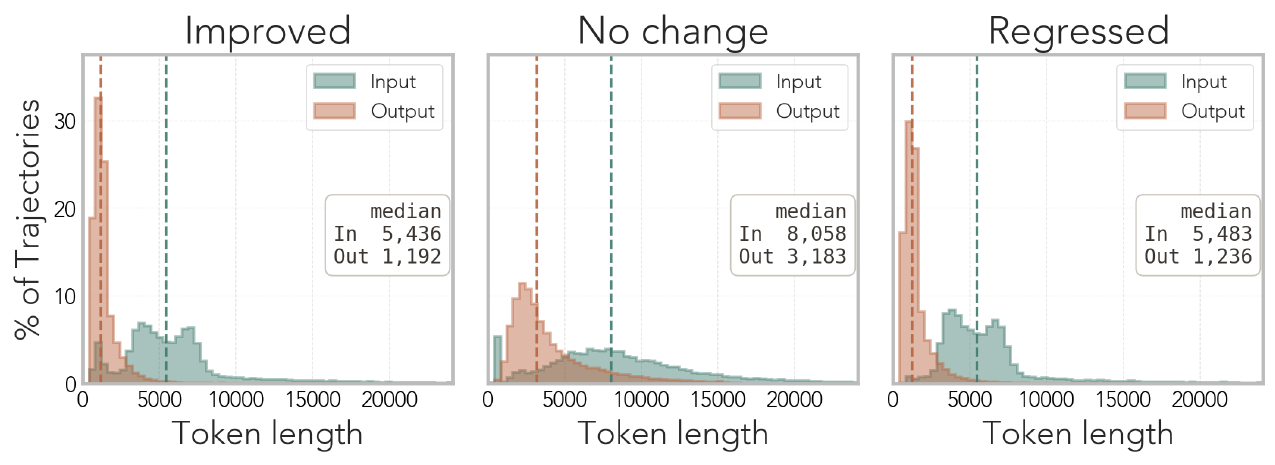}
    \vspace{0.2em}
    {\footnotesize (c) Trajectory length distribution by improvement type}

    \includegraphics[width=0.95\linewidth]{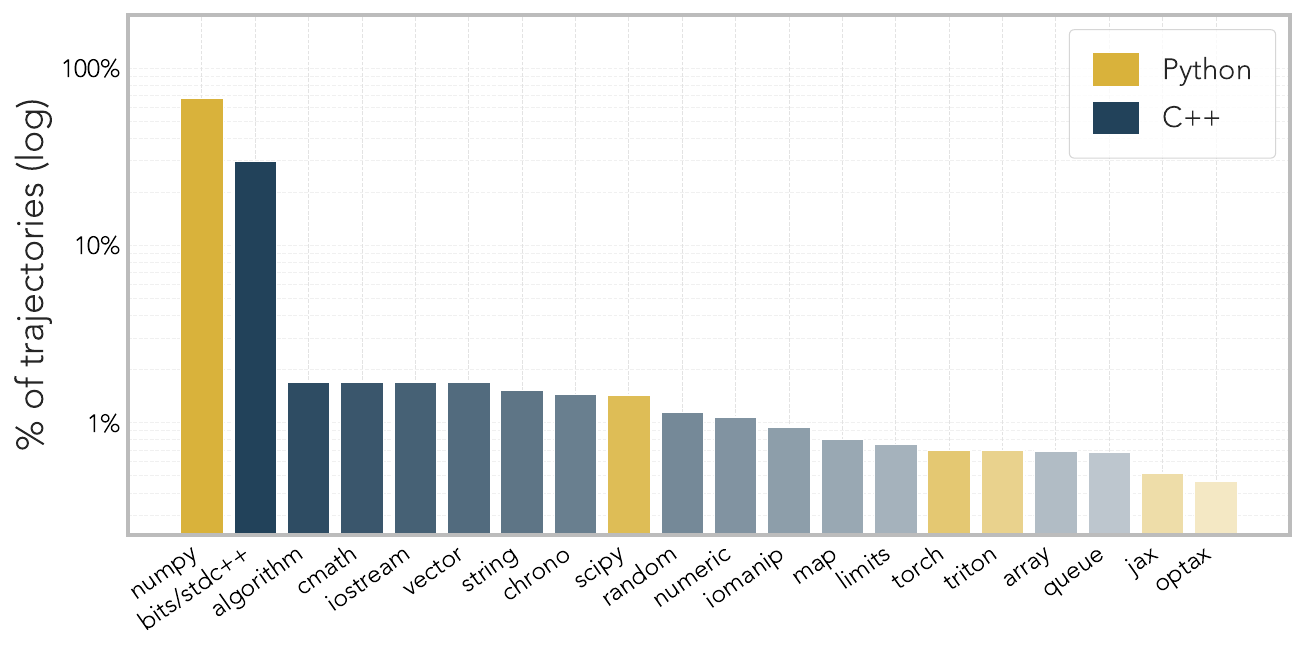}

    {\footnotesize (d) Top 20\% packages used in initial program}
\end{minipage}

\vspace{0.4em}
\caption{
Dataset composition and trajectory characteristics. 
(a) Number of tasks and trajectories for each task group. 
(b) Programming language and mutation strategy proportions across trajectories. 
(c) Distribution of trajectory lengths by improvement type (\improved, \nochange, and \regressed). 
(d) Top-20\% packages used in initial programs. 
}
\label{main_fig:dataset_statistics}
\end{figure}

\subsection{Analysis of \datasetName} \label{main_sec:dataset_analysis}

Overall, \datasetName contains 156,731 (\textasciitilde 156K) evolutionary search trajectories, covering 371 tasks across 10 task groups. As shown in Figure~\ref{main_fig:dataset_statistics} (a), Competitive Programming occupies the largest share with 172 individual tasks and is derived from FrontierCS~\citep{mang2025frontiercs}, followed by Numerical Algorithm Optimization, which builds upon ALE-Bench~\citep{imajuku2025ale}. 
In Figure~\ref{main_fig:dataset_statistics}~(b), 68.5\% of trajectories in \datasetName are based on Python, with a considerable portion in C++ (31.5\%). Regarding mutation strategy, \datasetName exhibits a well-balanced distribution between \diff (50.3\%) and \full (49.7\%). Together, these statistics suggest that models trained on \datasetName acquire optimization capabilities across both Python and C++, as well as proficiency in both \diff and \full strategies. Moreover, Figure~\ref{main_fig:dataset_statistics} (c) shows that the input length (avg. 8,902 tokens) is substantially longer than the output length (avg. 6,865 tokens, 1.3$\times$), indicating that \modelName learns to optimize parent programs by selectively leveraging useful feedback signals from a large context of prior programs. In addition, within \nochange, the median output length is 2.9$\times$ longer, suggesting that Qwen3.5-397B-A17B tends to engage in more extensive reasoning even when no improvement is achieved. Meanwhile, the output length in \regressed trajectories is comparable to that in \improved ones, implying that regressions arise not from insufficient reasoning, but from misguided exploration within the solution space. Moreover, in Figure~\ref{main_fig:dataset_statistics} (d), \datasetName most frequently utilizes \texttt{numpy}, followed by \texttt{bits/stdc++}, suggesting that the dataset spans both numerical computing and competitive programming workloads, thereby exposing models to diverse optimization patterns across multiple domains.

\subsection{\finchicon \modelName: Evolution Fine-Tuned Language Model} \label{main_sec:finch_model}

We introduce a new family of evolution fine-tuned LLMs, \finchicon \modelName, trained on \datasetName across 355 tasks, excluding 16 held-out tasks. As base models, we use the Qwen3.5~\citep{qwen35blog} series at three model sizes—2B, 4B, and 9B—as well as the Qwen3-8B~\citep{yang2025qwen3} for \modelName-8B. As shown in Figure~\ref{main_fig:task_group_dist}, the trajectory distribution is highly imbalanced, with Symbolic Regression (SR) trajectories constituting the majority, which could adversely affect training. To mitigate this, we use only one trajectory per task during training\footnote{Recall from Step 2 of dataset construction that three trajectories are collected per task.}. We fine-tune the base models using only \improved trajectories from \datasetName via full SFT, resulting in a total of 30,445 training trajectories. For validation, we use 900 trajectories uniformly sampled across tasks. For training, we employ the LLaMA-Factory~\citep{zheng2024llamafactory} framework. All models are trained for one epoch with a global batch size of 128 and a learning rate of $1\text{e-}5$. All experiments are conducted on eight NVIDIA H200 140GB GPUs.

\section{Experiments} \label{main_sec:expr}

\subsection{Experimental Setup}

\begin{table}[!t]
\centering
\small
\setlength{\tabcolsep}{3.5pt}

\caption{Performance comparison of \finchicon \modelName to base models across Mathematical Discovery, Algorithm Engineering, and System Performance benchmarks. \textsuperscript{\dag} Scores reported in \citep{qu2026coral}. \textsuperscript{\ddag} Scores reported in \citep{liu2026evox}. $\Delta$ denotes the relative improvement (\%) of \modelName over the corresponding same-size base model, sign-adjusted so that positive values always indicate improvement regardless of metric direction. \textbf{Avg.\ Gain} is the arithmetic mean of available $\Delta$ values, excluding \textbf{ahc058} where the base score is near zero and yields a disproportionately large ratio.}

\vspace{5pt}
\label{main_tab:eft_test_time_search_results}

\renewcommand{\arraystretch}{1.15}

\begin{adjustbox}{max width=\linewidth}
\begin{tabular}{lcccccccccccc}
\toprule

& \multicolumn{5}{c}{\textbf{Mathematical Discovery}}
& \multicolumn{2}{c}{\textbf{Algorithm Engineering}}
& \multicolumn{4}{c}{\textbf{System Performance}} & \\

\cmidrule(lr){2-6}
\cmidrule(lr){7-8}
\cmidrule(lr){9-12}

\textbf{Model}
& \textbf{Erd\H{o}s}($\downarrow$)
& \textbf{AC1}($\downarrow$)
& \textbf{AC2}($\uparrow$)
& \textbf{CP(n=26)}($\uparrow$)
& \textbf{Hadamard}($\uparrow$)
& \textbf{ahc039}($\uparrow$)
& \textbf{ahc058}($\uparrow$)
& \textbf{EPLB}($\uparrow$)
& \textbf{PRISM}($\uparrow$)
& \textbf{LLM-SQL}($\uparrow$)
& \textbf{Transaction}($\uparrow$)
& \textbf{Avg.}($\uparrow$) \\

\midrule

Best Human
& 0.380927
& 1.5097
& 0.9015
& 2.634000
& 0.935673
& 566{,}997
& 847{,}674{,}723
& 0.1265
& 21.89
& 0.6920
& 2724.80
& -- \\

\hdashline

Initial Program
& 0.495056
& 1.5186
& 0.8558
& 0.959764
& 0.143275
& 534{,}850
& 0
& 0.1265
& 21.89
& 0.6856
& 2824.86
& -- \\

\midrule

\multicolumn{13}{c}{\textbf{OpenEvolve + Proprietary Models}} \\

\midrule

Claude-Opus-4.6~\textsuperscript{\dag}
& 0.381880
& --
& --
& 2.629300
& --
& --
& --
& 0.1270
& 26.26
& 0.7160
& 3774.00
& -- \\

Gemini-3-Pro~\textsuperscript{\ddag}
& --
& --
& --
& 2.541400
& --
& --
& --
& 0.1272
& 26.24
& 0.7258
& 4273.50
& -- \\

GPT-5~\textsuperscript{\ddag}
& --
& --
& --
& 2.541400
& --
& --
& --
& 0.1272
& 26.23
& 0.7155
& 4237.30
& -- \\

\midrule

\multicolumn{13}{c}{\textbf{OpenEvolve + Open-source Models}} \\

\midrule
Qwen3.5-2B
& 0.381737
& 1.5186
& 0.8646
& 1.253056
& 0.478009
& 546{,}078
& 0
& 0.1265
& 21.89
& 0.6856
& 2832.86
& -- \\

\finchcolor \modelName-2B
& 0.381346
& 1.5184
& 0.8920
& 1.535134
& 0.400476
& 545{,}256
& 329{,}359{,}253
& 0.1269
& 22.26
& 0.6860
& 2949.85
& -- \\

\deltarowcolor $\Delta$
& \good{+0.10\%}
& \good{+0.01\%}
& \good{+3.17\%}
& \good{+22.51\%}
& \bad{-16.22\%}
& \bad{-0.15\%}
& --
& \good{+0.32\%}
& \good{+1.69\%}
& \good{+0.06\%}
& \good{+4.13\%}
& \good{\textbf{+1.56\%}} \\

\midrule

Qwen3.5-4B
& 0.416924
& 1.5186
& 0.8802
& 1.680787
& 0.384332
& 542{,}077
& 0
& 0.1266
& 21.89
& 0.6856
& 2732.24
& -- \\

\finchcolor \modelName-4B
& 0.386460
& 1.5173
& 0.8933
& 1.806808
& 0.146199
& 551{,}844
& 331{,}466{,}883
& 0.1267
& 22.87
& 0.6857
& 4761.90
& -- \\

\deltarowcolor $\Delta$
& \good{+7.31\%}
& \good{+0.09\%}
& \good{+1.49\%}
& \good{+7.50\%}
& \bad{-61.96\%}
& \good{+1.80\%}
& --
& \good{+0.08\%}
& \good{+4.48\%}
& \good{+0.01\%}
& \good{+74.30\%}
& \good{\textbf{+3.40\%}} \\

\midrule

Qwen3-8B
& 0.403585
& 1.5177
& 0.8980
& 1.797576
& 0.452330
& 557{,}081
& 0
& 0.1269
& 23.81
& 0.6858
& 3174.60
& -- \\

\finchcolor \modelName-8B
& 0.381236
& 1.5154
& 0.9001
& 1.822617
& 0.501743
& 557{,}168
& 135{,}184{,}684
& 0.1270
& 24.70
& 0.7341
& 3257.33
& -- \\

\deltarowcolor $\Delta$
& \good{+5.54\%}
& \good{+0.15\%}
& \good{+0.23\%}
& \good{+1.39\%}
& \good{+10.92\%}
& \good{+0.02\%}
& --
& \good{+0.08\%}
& \good{+3.74\%}
& \good{+7.04\%}
& \good{+2.61\%}
& \good{\textbf{+3.17\%}} \\

\midrule

Qwen3.5-9B
& 0.385512
& 1.5186
& 0.8801
& 1.172702
& 0.397184
& 553{,}582
& 134{,}486{,}700
& 0.1269
& 22.36
& 0.6858
& 3584.23
& -- \\

\finchcolor \modelName-9B
& 0.381100
& 1.5141
& 0.9122
& 1.936000
& 0.480585
& 553{,}759
& 525{,}286{,}896
& 0.1265
& 23.93
& 0.7024
& 3636.36
& -- \\

\deltarowcolor $\Delta$
& \good{+1.14\%}
& \good{+0.30\%}
& \good{+3.65\%}
& \good{+65.09\%}
& \good{+21.00\%}
& \good{+0.03\%}
& \good{+290.59\%}
& \bad{-0.32\%}
& \good{+7.02\%}
& \good{+2.42\%}
& \good{+1.45\%}
& \good{\textbf{+10.24\%}} \\

\bottomrule
\end{tabular}
\end{adjustbox}
\end{table}

\paragraph{Evaluation Tasks.}

To evaluate the \textbf{cross-task discovery generalization} of \modelName, we use a diverse set of optimization tasks that are, to the extent possible, disjoint from those used to train \modelName. These tasks are drawn from benchmarks widely adopted in prior works~\citep{cemri2026adaevolve,liu2026evox,skydiscover2026,yuksekgonul2026learning,ye2026evaluation}. In total, we evaluate \modelName across five domains and 22 tasks: \textbf{(1) Mathematical Discovery}, including the Erdős Minimum Overlap Problem (Erdos), First Autocorrelation Inequality (AC1), Second Autocorrelation Inequality (AC2), Circle Packing in a Unit Square with $n=26$ (CP), and Hadamard Maximum Determinant (Hadamard); \textbf{(2) Algorithm Engineering}, including two tasks, ahc039 and ahc048; \textbf{(3) System Performance}, including four tasks, EPLB, PRISM, LLM-SQL, and Transaction; \textbf{(4) Competitive Programming}, comprising six tasks, each scored on a scale from 0 to 100. These include Problem 263 (P263) from CALICO\footnote{\url{https://frontier-cs.org/blog/calico/}}, UC Berkeley's official programming contest featuring open-ended optimization problems, and five newly added FrontierCS v1.1 problems, Problem 301--305 (P301--305); and \textbf{(5) Algorithmic Heuristics}, including five tasks, Convolve2D Full Fill (Convolve2D), PolynomialReal (Polynomial), Positive Semidefinite Cone Projection (PSD), 2D Affine Transform (Affine Transform), and FFT Convolution (FFT Conv.). Although some tasks are drawn from the same benchmark suite, we treat them as independent evaluation tasks because discovery tasks often differ substantially in their objectives, search spaces, and solution forms. For evaluation, following the definition of discovery, we report the maximum task-specific score achieved within $T$ iterations. Each optimization task uses its own task-specific metric, such as the \texttt{c5\_bound} for the Erdős overlap problem. Detailed descriptions of these task-specific metrics are provided in Appendix~\ref{supp_sec:full_task_list}.

\paragraph{Baselines.}

We evaluate \modelName's capability as a mutation operator when combined with a test-time search scaffold; throughout this work, we use OpenEvolve~\citep{openevolve} as the default scaffold unless otherwise noted. Specifically, we measure the relative improvement of \modelName over the initial program score and compare it against two baselines: (1) the base model with a search scaffold; and (2) the base model with a learning scaffold (\eg TTT-Discover~\citep{yuksekgonul2026learning}). However, since running the original TTT-Discover is prohibitively expensive---each task requires up to 50 epochs to reproduce, costing approximately 500 USD on average---we instead adopt nanodiscover\footnote{\url{https://github.com/cheongalc/nanodiscover/}}, an open-source reproduction of TTT-Discover that does not depend on the Tinker API. A detailed description of nanodiscover is provided in Appendix~\ref{supp_sec:additional_implementation_details}.

\paragraph{Inference Details.}

For the search-based scaffold, following prior work~\citep{liu2026evox}, we set $T=100$ with a parallel evaluation size of 1, temperature 0.7, top-$p$ 0.95, and a maximum of 30K tokens. For the remaining scaffold-specific hyperparameters, such as the island size, we adopt the default settings specified for each task. For the learning-based scaffold, we follow the same configurations as the original TTT-Discover, which is presented in Appendix~\ref{supp_sec:additional_implementation_details}.

\subsection{Impact of Evolution Fine-Tuning} \label{main_sec:expr_results}

\begin{figure*}[t!]
  \centering
  \begin{minipage}[t]{0.66\textwidth}
    \centering
    \captionof{table}{Competitive Programming performance of \finchicon \modelName compared to base models across six optimization tasks.}
    \label{main_tab:frontiercs_results}
    \renewcommand{\arraystretch}{1.2}
    \begin{adjustbox}{max width=\linewidth}
    \begin{tabular}{lccccccc}
    \toprule
    \textbf{Model} & \textbf{P263} ($\uparrow$) & \textbf{P301} ($\uparrow$) & \textbf{P302} ($\uparrow$) & \textbf{P303} ($\uparrow$) & \textbf{P304} ($\uparrow$) & \textbf{P305} ($\uparrow$) & \textbf{Avg.} ($\uparrow$) \\
    \midrule
    Qwen3.5-2B                   & 0.00  & 0.00  & 0.00  & 0.00  & 0.00  & 0.00  & 0.00 \\
    \finchcolor \modelName-2B    & 0.38  & 1.63  & 0.12  & 3.16  & 0.00  & 31.43 & \textbf{6.12} \\
    \midrule
    Qwen3.5-4B                   & 8.15  & 20.99 & 27.07 & 0.00  & 0.00  & 30.89 & 14.52 \\
    \finchcolor \modelName-4B    & 27.44 & 68.17 & 24.41 & 31.79 & 0.00  & 40.03 & \textbf{31.97} \\
    \midrule
    Qwen3-8B                     & 23.72 & 44.67 & 36.78 & 10.84 & 0.00  & 29.23 & 24.21 \\
    \finchcolor \modelName-8B    & 38.12 & 39.41 & 36.68 & 10.84 & 0.00  & 22.34 & \textbf{24.56} \\
    \midrule
    Qwen3.5-9B                   & 55.09 & 27.59 & 35.63 & 35.54 & 5.81  & 35.14 & 32.46 \\
    \finchcolor \modelName-9B    & 86.10 & 58.78 & 36.68 & 34.02 & 22.11 & 38.38 & \textbf{46.01} \\
    \bottomrule
    \end{tabular}
    \end{adjustbox}
  \end{minipage}
  \hfill
  \begin{minipage}[t]{0.33\textwidth}
    \centering
    \vspace{0pt}
    \includegraphics[width=\linewidth]{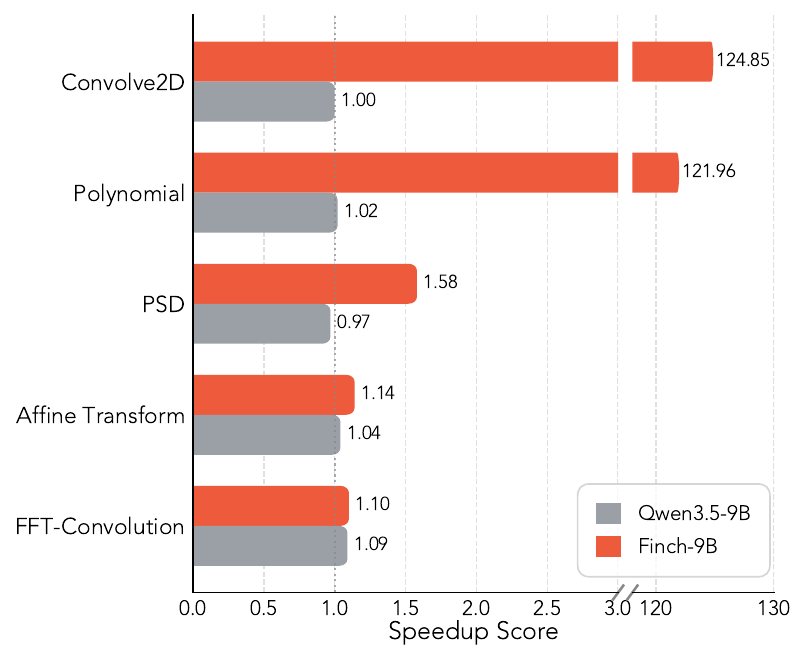}
    \captionof{figure}{Algorithmic Heuristics performance of \finchicon \modelName compared to Qwen3.5-9B.}
    \label{main_fig:algotune_results}
  \end{minipage}
\end{figure*}

\paragraph{EFT improves cross-task discovery generalization across all model scales.}
Table~\ref{main_tab:eft_test_time_search_results} shows that \modelName, evolution fine-tuned on \datasetName, achieves substantial relative performance gains across 22 tasks, with the largest improvements observed on ahc058 (+290.59\%) and Transaction (+74.30\%). In Table~\ref{main_tab:frontiercs_results} and Figure~\ref{main_fig:algotune_results}, we observe that EFT is also effective on complex optimization-intensive tasks, including NP-hard competitive programming and algorithmic heuristics. Moreover, larger models benefit more from \datasetName: \modelName-9B obtains a larger relative gain (+10.31\%) than \modelName-4B (+3.40\%). Our findings further demonstrate that EFT enables smaller models to match or even exceed the performance of non-EFT models twice their size; for instance, \modelName-4B achieves 0.386460 on Erdos, comparable to Qwen3-8B's 0.403585. Taken together, these results suggest that \datasetName enables LLMs to internalize discovery capabilities, thereby exhibiting cross-task discovery generalization.

\begin{figure*}[t!]
  \centering
  \begin{minipage}[t]{0.42\textwidth}
    \setlength{\tabcolsep}{4pt}
    \renewcommand{\arraystretch}{1.1}
    \centering
    \captionof{table}{Effect of Offline RL (CP: avg.\ competitive programming score).}
    \label{main_tab:eft_kto_results}
    \begin{adjustbox}{width=0.95\textwidth}
    \begin{tabular}{lcccc}
    \toprule
    \textbf{Model} & \textbf{Erd\H{o}s} ($\downarrow$) & \textbf{AC1} ($\downarrow$) & \textbf{AC2} ($\uparrow$) & \textbf{CP} ($\uparrow$) \\
    \midrule
    Best Human & \textbf{0.380927} & 1.5097 & 0.9015 & -- \\
    \hdashline
    Qwen3.5-4B & 0.416924 & 1.5186 & 0.8802 & 14.52 \\
    \finchcolor \modelName-4B & 0.386460 & 1.5173 & 0.8933 & 31.97 \\
    \finchcolor \modelName-4B + KTO & 0.381809 & 1.5151 & 0.9121 & \textbf{36.30} \\
    \midrule
    Qwen3-8B & 0.403585 & 1.5177 & 0.8980 & 24.21 \\
    \finchcolor \modelName-8B & \textbf{0.381236} & 1.5154 & 0.9001 & 24.56 \\
    \finchcolor \modelName-8B + KTO & 0.381596 & \textbf{1.5089} & \textbf{0.9146} & \textbf{37.30} \\
    \bottomrule
    \end{tabular}
    \end{adjustbox}
  \end{minipage}
  \hfill
  \begin{minipage}[t]{0.56\textwidth}
    \setlength{\tabcolsep}{4pt}
    \renewcommand{\arraystretch}{1.2}
    \centering
    \captionof{table}{Performance of \finchicon \modelName combined with online (test-time) RL scaffolds across math optimization tasks.}
    \label{main_tab:eft_test_time_rl_results}
    \begin{adjustbox}{max width=\linewidth}
    \begin{tabular}{@{}ll|ccc@{}}
    \toprule
    \textbf{Scaffold} & \textbf{Model}
      & \shortstack{\textbf{Erd\H{o}s} ($\downarrow$)}
      & \shortstack{\textbf{CP (n=26)} ($\uparrow$)}
      & \shortstack{\textbf{CP (n=32)} ($\uparrow$)} \\
    \midrule
    ThetaEvolve & R1-Qwen3-8B
      & -
      & \textbf{2.635983} & - \\
    \midrule
    TTT-Discover & GPT-OSS-120B
      & 0.380876
      & - & - \\
                 & Qwen3-8B
      & \textbf{0.380932}
      & \textbf{2.635983} & 2.939572 \\
    \midrule
    nanodiscover & Qwen3-8B
      & 0.380956
      & \textbf{2.635983} & \textbf{2.939573} \\
                \finchcolor  & \modelName-8B
      & \textbf{0.380948}
      & \textbf{2.635983} & \textbf{2.939573} \\
    \bottomrule
    \end{tabular}
    \end{adjustbox}
  \end{minipage}
\end{figure*}

\paragraph{Teaching \modelName to distinguish good from bad solutions further enhances cross-task discovery generalization.}
After EFT, we further train \modelName using preference learning (KTO~\citep{ethayarajh2024kto}) on \improved and \regressed jointly, in order to examine whether discovery capability can be further internalized by enabling \modelName to self-judge which solutions are good and which are not. As shown in Table~\ref{main_tab:eft_kto_results}, offline RL consistently improves \modelName's discovery capability; notably, \modelName-8B with KTO surpasses the best human score on both the AC1 and AC2 tasks. These results suggest that (1) \datasetName provides a useful and complementary training signal beyond supervised fine-tuning, and (2) internalizing the ability to discriminate good from bad solutions is a viable path toward instilling discovery skills directly into the model's parameters, rather than relying solely on test-time search.

\begin{figure*}[t]
\centering

\begin{minipage}[t]{0.62\textwidth}
    \centering
    \vspace{0pt}
    \includegraphics[width=\linewidth]{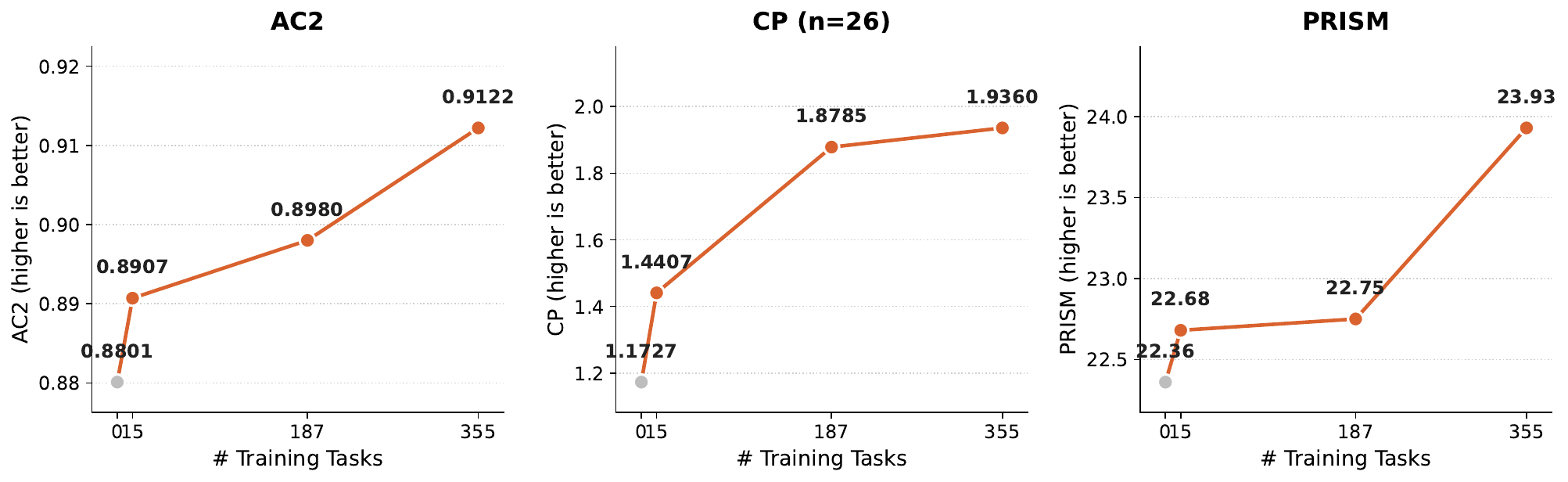}
    \captionof{figure}{Scaling trends with increasing numbers of training tasks in \datasetName, evaluated on AC2, CP ($n=26$), and PRISM.}
    \label{main_fig:scaling_plot}
\end{minipage}
\hfill
\begin{minipage}[t]{0.35\textwidth}
    \centering
    \vspace{0pt}
    \captionof{table}{Effect of improvement type on EFT (CP: avg. competitive programming score)}
    \label{main_tab:hardsoft1}
    \resizebox{\linewidth}{!}{
    \begin{tabular}{lccc}
    \toprule
    \textbf{Model} & \textbf{Erd\H{o}s} ($\downarrow$) & \textbf{AC2} ($\uparrow$) & \textbf{CP} ($\uparrow$) \\
    \midrule
    Best Human & 0.380927 & 0.9015 & - \\
    \hdashline
    Qwen3-8B & 0.403585 & 0.8980 & 24.21 \\
    + \improved & 0.381236 & 0.9001 & 24.56 \\
    + \improved + \nochange & 0.381261 & 0.9052 & 19.56 \\
    \midrule
    Qwen3.5-9B & 0.385512 & 0.8801 & 32.46 \\
    + \improved & \textbf{0.381100} & 0.9122 & 46.01 \\
    + \improved + \nochange & 0.399486 & \textbf{0.9148} & \textbf{50.12} \\
    \bottomrule
    \end{tabular}
    }
\end{minipage}

\end{figure*}

\paragraph{EFT serves as mid-training for test-time RL.}
We apply test-time RL to \modelName using nanodiscover. Table~\ref{main_tab:eft_test_time_rl_results} shows that \modelName achieves the best performance on two circle-packing tests and improves performance on Erd\H{o}s (+3.2\%). These results suggest that EFT can serve as a form of mid-training that strengthens test-time RL. However, compared with the original TTT-Discover using GPT-OSS-120B, \modelName still achieves lower performance, indicating that it remains difficult for smaller models (\eg 8B-scale) to discover push-frontier solutions.

\section{Analysis} \label{main_sec:analysis}

\begin{wrapfigure}[11]{r}{0.45\textwidth}
    \centering
    \vspace{-5em}
    \includegraphics[width=0.5\textwidth]{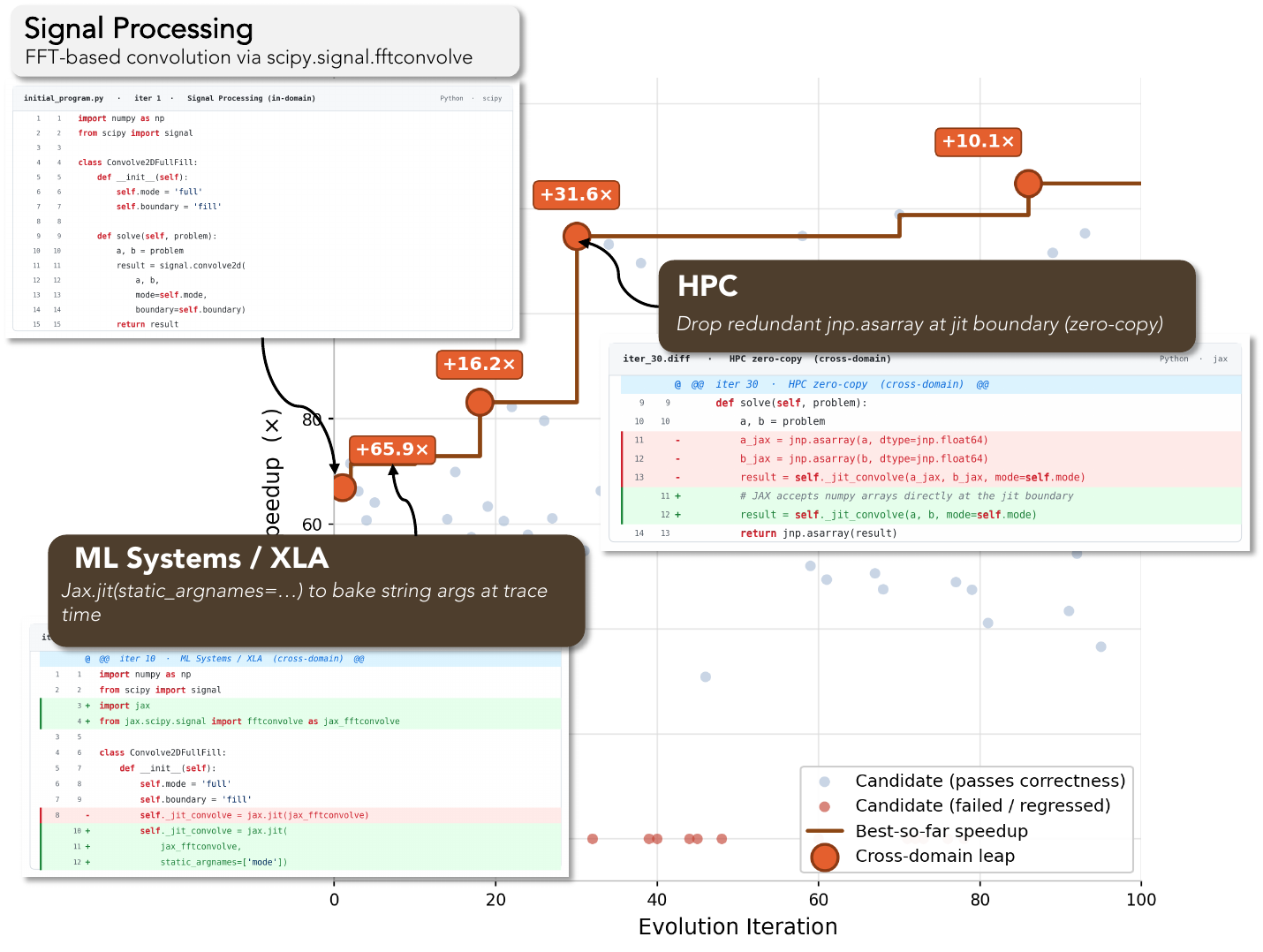}
    \captionsetup{justification=centering}
    \caption{Case Study on Convolve2D.}
    \label{main_fig:case_study}
\end{wrapfigure}

\paragraph{Scaling the number of tasks improves cross-task discovery transfer.}
As shown in Figure~\ref{main_fig:scaling_plot}, performance on AC2, CP ($n=26$), and PRISM exhibits a clear positive scaling trend as the number of training tasks in \datasetName increases. These results indicate that \datasetName provides a scalable training signal: increasing the amount of \datasetName consistently improves discovery performance. This suggests that the gains from EFT are not merely task-specific, but can further grow with larger task collections, more trajectories, and increased model capacity.

\paragraph{\modelName can transfer discovery pattern by adopting different domain knowledge.}

Figure~\ref{main_fig:case_study} presents a case study on the Convolve2D task. In this example, \modelName-9B modifies the implementation from \texttt{scipy} to \texttt{jax} to improve computational efficiency. We hypothesize that this behavior emerges because \datasetName contains a substantial number of trajectories involving the \texttt{jax} library, particularly from tasks such as uncertainty inequalities and matrix multiplication, as shown in Figure~\ref{main_fig:dataset_statistics} (d). These results suggest that \datasetName enables LLMs to internalize transferable discovery patterns across domains.

\section{Related Work} \label{main_sec:related_work}

\paragraph{LLM-driven Evolutionary Search Scaffolds.}
Recent work shows that LLM-driven evolutionary search scaffolds can produce novel solutions across diverse optimization tasks. These methods can be broadly categorized into \emph{search-based} and \emph{learning-based} approaches. Search-based scaffolds primarily differ in how they archive candidate solutions and select parents for mutation. For instance, AlphaEvolve~\citep{novikov2025alphaevolve} employs MAP-Elites with island-based populations to balance exploration and exploitation, while OpenEvolve adopts a similar MAP-Elites framework with periodic migration. CodeEvolve~\citep{assumpccao2025codeevolve} integrates island-based genetic algorithms with inspiration-based crossover and meta-prompting, and ShinkaEvolve~\citep{lange2025shinkaevolve} replaces fixed quality–diversity grids with a sample-efficient regime combining weighted sampling, novelty rejection, and bandit-based LLM selection. Other methods explore alternative selection and adaptation strategies: GEPA operates along Pareto frontiers; PACEvolve~\citep{yan2026pacevolve} and AdaEvolve~\citep{cemri2026adaevolve} emphasize long-horizon, progress-aware, and adaptive proposal mechanisms; and EvoX~\citep{liu2026evox} and SkyDiscover~\citep{skydiscover2026} study meta-evolution and unified discovery frameworks. Application-specific extensions include KernelEvolve~\citep{liao2025kernelevolve} and Kernel-Smith~\citep{du2026kernel} for GPU kernel optimization, while the Darwin Gödel Machine~\citep{zhang2025darwin}, HyperAgents~\citep{zhang2026hyperagents}, and CORAL~\citep{qu2026coral} investigate open-ended self-improvement and multi-agent evolution. Meta-Harness~\citep{lee2026meta} further shifts the optimization target from candidate solutions to the \emph{harness} itself. In contrast, learning-based approaches such as ThetaEvolve~\citep{wang2025thetaevolve} and TTT-Discover~\citep{yuksekgonul2026learning} demonstrate that combining evolutionary search with test-time learning enables LLMs to internalize discovery capabilities.

\paragraph{Benchmarks for Optimization and Scientific Discovery.}
Progress in LLM-driven discovery has been driven by benchmarks that emphasize long-horizon, open-ended problem solving under strong baselines. KernelBench~\citep{ouyang2025kernelbench} and GSO~\citep{shetty2025gso} evaluate iterative optimization of GPU kernels and programs, while AlgoTune~\citep{press2025algotune} measures performance gains in numerical routines. ALE-Bench~\citep{imajuku2025ale} focuses on long-horizon algorithm engineering tasks derived from programming contests. FrontierCS~\citep{mang2025frontiercs} and AutoLab~\citep{autolab-2026} introduce \emph{living} benchmarks that evolve alongside frontier models and support end-to-end scientific–engineering loops. For scientific-law discovery, LLM-SRBench~\citep{shojaee2025llm} evaluates whether models can recover symbolic equations from data, while \citep{lin2025can} examine the rediscovery of empirical scaling laws.

\section{Conclusion} \label{main_sec:conclusion}

In this work, we show that evolution fine-tuning (EFT) — fine-tuning LLMs on a large collection of evolutionary trajectories spanning 371 tasks — improves the model's discovery capability on 22 held-out tasks, demonstrating cross-task discovery generalization. Furthermore, we show that \modelName exhibits a synergistic effect when combined with test-time RL.

\section*{Limitations} \label{main_sec:limitation}

\paragraph{Mixed Test-Time Search Scaffolds.}
In this work, we use only the OpenEvolve scaffold for both collecting trajectories and evaluating \modelName. A concern is that a model trained on OpenEvolve-style trajectories may not generalize well when combined with stronger scaffolds, such as EvoX~\citep{liu2026evox}. To mitigate this issue in future work, collecting trajectories from mixed scaffolds will be necessary as a form of template input variation~\citep{longpre2023flan} to improve cross-scaffold generalization.

\paragraph{Extending Test-Time RL Experiments to Diverse and Realistic Tasks.}
In this work, we demonstrate that \modelName exhibits a positive synergistic effect with test-time RL only on mathematical tasks. However, many practical tasks exist in the real world, such as kernel engineering. It is therefore important to verify whether our model also exhibits a positive synergistic effect with test-time RL on these tasks in order to establish the broader effectiveness of our approach.

\paragraph{Beyond Language Modality and Single-turn.}

Scientific discovery is not limited to the language modality alone. In many scientific domains, researchers must also interpret visual observations from experiments~\citep{sun2025scienceboard,ma2026orion}, making multi-modal reasoning an essential component of the discovery process. Similar to how we internalize discovery capabilities into LLMs in this work, our training recipe can be naturally extended to the multi-modal setting. In particular, recent advances in distillation techniques~\citep{lee2024collavo,lee2024moai,lee2024meteor,lee2024phantom,lee2024trol,lee2025genrecal,lee2025vlsi,lee2026unified,kang2026agent,lee2026masking,lee2026recursive,cho2026spatialclaw,yu2026hide,kim2026and} for Vision--Language Models are highly complementary to our approach and could enable \textsc{Evolution Fine-Tuning} for multi-modal scientific discovery. Furthermore, our current framework trains the model to generate a child solution conditioned on the evolutionary history and parent program in a single turn. An important future direction is to extend this paradigm to multi-turn interactions~\citep{lee2024dialogcc,lee2024stark,lee2024thanos,lee2024large,lee2025multiverse,lee2025refinebench}, allowing the model to continually reason over previously explored lineages and iteratively learn to evolve solutions in more promising directions.

\section*{Acknowledgement}

This research was supported by the ``Advanced GPU Utilization Support Program'' funded by the Government of the Republic of Korea (Ministry of Science and ICT). We thank the SkyDiscover team for their valuable feedback on the dataset construction process, the use of the SkyDiscover framework, and the overall direction of this research. In particular, we would like to thank Shu Liu, Shubham Agarwal, and Mert Cemri for their insightful comments and discussions. We also thank the OpenEvolve team, especially Ritik Vijayvergiya and Asankhaya Sharma, for their helpful feedback on the use of the OpenEvolve framework and for their thoughtful comments on this work. We are also grateful to the authors of ALE-Bench, especially Yuki Imajuku, and the AtCoder team for authorizing the public release of the evolutionary search trajectories derived from their CC BY-ND 4.0 licensed dataset. We thank Minnesota NLP group members for valuable feedback. In addition, we thank Byung-Kwan Lee, a Research Scientist at NVIDIA, for providing valuable feedback during the early stages of this project. Finally, we sincerely thank Tahee Jung from Amazon for providing extensive and valuable feedback throughout this project.

{\small
\bibliographystyle{unsrtnat}
\bibliography{neurips_2026}

@article{novikov2025alphaevolve,
  title={Alphaevolve: A coding agent for scientific and algorithmic discovery},
  author={Novikov, Alexander and V{\~u}, Ng{\^a}n and Eisenberger, Marvin and Dupont, Emilien and Huang, Po-Sen and Wagner, Adam Zsolt and Shirobokov, Sergey and Kozlovskii, Borislav and Ruiz, Francisco JR and Mehrabian, Abbas and others},
  journal={arXiv preprint arXiv:2506.13131},
  year={2025}
}

@article{romera2024mathematical,
    title     = {Mathematical discoveries from program search with large language models},
    author    = {Romera-Paredes, Bernardino and Barekatain, Mohammadamin and Novikov, Alexander and Balog, Matej and Kumar, M Pawan and Dupont, Emilien and Ruiz, Francisco JR and Ellenberg, Jordan S and Wang, Pengming and Fawzi, Omar and others},
    journal   = {Nature},
    volume    = {625},
    number    = {7995},
    pages     = {468--475},
    year      = {2024},
    publisher = {Nature Publishing Group UK London}
}

@article{yang2025qwen3,
  title={Qwen3 technical report},
  author={Yang, An and Li, Anfeng and Yang, Baosong and Zhang, Beichen and Hui, Binyuan and Zheng, Bo and Yu, Bowen and Gao, Chang and Huang, Chengen and Lv, Chenxu and others},
  journal={arXiv preprint arXiv:2505.09388},
  year={2025}
}

@article{ethayarajh2024kto,
  title={Kto: Model alignment as prospect theoretic optimization},
  author={Ethayarajh, Kawin and Xu, Winnie and Muennighoff, Niklas and Jurafsky, Dan and Kiela, Douwe},
  journal={arXiv preprint arXiv:2402.01306},
  year={2024}
}

@article{zhang2025darwin,
  title={Darwin godel machine: Open-ended evolution of self-improving agents},
  author={Zhang, Jenny and Hu, Shengran and Lu, Cong and Lange, Robert and Clune, Jeff},
  journal={arXiv preprint arXiv:2505.22954},
  year={2025}
}

@article{zhang2026hyperagents,
  title={Hyperagents},
  author={Zhang, Jenny and Zhao, Bingchen and Yang, Wannan and Foerster, Jakob and Clune, Jeff and Jiang, Minqi and Devlin, Sam and Shavrina, Tatiana},
  journal={arXiv preprint arXiv:2603.19461},
  year={2026}
}

@misc{openevolve,
  author       = {Sharma, Asankhaya},
  title        = {{OpenEvolve}: An Open-Source Evolutionary Coding Agent},
  year         = {2025},
  howpublished = {\url{https://github.com/algorithmicsuperintelligence/openevolve}},
  note         = {GitHub repository}
}

@article{lange2025shinkaevolve,
  title={Shinkaevolve: Towards open-ended and sample-efficient program evolution},
  author={Lange, Robert Tjarko and Imajuku, Yuki and Cetin, Edoardo},
  journal={arXiv preprint arXiv:2509.19349},
  year={2025}
}

@article{assumpccao2025codeevolve,
  title={Codeevolve: An open source evolutionary coding agent for algorithm discovery and optimization},
  author={Assump{\c{c}}{\~a}o, Henrique and Ferreira, Diego and Campos, Leandro and Murai, Fabricio},
  journal={arXiv preprint arXiv:2510.14150},
  year={2025}
}

@article{agrawal2025gepa,
  title={Gepa: Reflective prompt evolution can outperform reinforcement learning},
  author={Agrawal, Lakshya A and Tan, Shangyin and Soylu, Dilara and Ziems, Noah and Khare, Rishi and Opsahl-Ong, Krista and Singhvi, Arnav and Shandilya, Herumb and Ryan, Michael J and Jiang, Meng and others},
  journal={arXiv preprint arXiv:2507.19457},
  year={2025}
}

@article{lee2026meta,
  title={Meta-Harness: End-to-End Optimization of Model Harnesses},
  author={Lee, Yoonho and Nair, Roshen and Zhang, Qizheng and Lee, Kangwook and Khattab, Omar and Finn, Chelsea},
  journal={arXiv preprint arXiv:2603.28052},
  year={2026}
}

@article{yan2026pacevolve,
  title={Pacevolve: Enabling long-horizon progress-aware consistent evolution},
  author={Yan, Minghao and Peng, Bo and Coleman, Benjamin and Chen, Ziqi and Xie, Zhouhang and Chen, Shuo and He, Zhankui and Sachdeva, Noveen and Ye, Isabella and Wang, Weili and others},
  journal={arXiv preprint arXiv:2601.10657},
  year={2026}
}

@article{cemri2026adaevolve,
  title={AdaEvolve: Adaptive LLM driven zeroth-order optimization},
  author={Cemri, Mert and Agrawal, Shubham and Gupta, Akshat and Liu, Shu and Cheng, Audrey and Mang, Qiuyang and Naren, Ashwin and Erdogan, Lutfi Eren and Sen, Koushik and Zaharia, Matei and others},
  journal={arXiv preprint arXiv:2602.20133},
  year={2026}
}

@article{qu2026coral,
  title={CORAL: Towards Autonomous Multi-Agent Evolution for Open-Ended Discovery},
  author={Qu, Ao and Zheng, Han and Zhou, Zijian and Yan, Yihao and Tang, Yihong and Ong, Shao Yong and Hong, Fenglu and Zhou, Kaichen and Jiang, Chonghe and Kong, Minwei and others},
  journal={arXiv preprint arXiv:2604.01658},
  year={2026}
}

@article{liu2026evox,
  title={EvoX: Meta-Evolution for Automated Discovery},
  author={Liu, Shu and Agarwal, Shubham and Maheswaran, Monishwaran and Cemri, Mert and Li, Zhifei and Mang, Qiuyang and Naren, Ashwin and Boneh, Ethan and Cheng, Audrey and Pan, Melissa Z and others},
  journal={arXiv preprint arXiv:2602.23413},
  year={2026}
}

@misc{skydiscover2026,
  title = {SkyDiscover: A Flexible Framework for AI-Driven Scientific and Algorithmic Discovery},
  author = {Liu, Shu and Cemri, Mert and Agarwal, Shubham and Krentsel, Alexander and Naren, Ashwin and Mang, Qiuyang and Li, Zhifei and Gupta, Akshat and Maheswaran, Monishwaran and Cheng, Audrey and Pan, Melissa and Boneh, Ethan and Ramchandran, Kannan and Sen, Koushik and Dimakis, Alexandros G. and Zaharia, Matei and Stoica, Ion},
  year = {2026},
  url = {https://skydiscover-ai.github.io/blog.html}
}

@article{liao2025kernelevolve,
  title={Kernelevolve: Scaling agentic kernel coding for heterogeneous ai accelerators at meta},
  author={Liao, Gang and Qin, Hongsen and Wang, Ying and Golden, Alicia and Kuchnik, Michael and Yetim, Yavuz and Ang, Jia Jiunn and Fu, Chunli and He, Yihan and Hsia, Samuel and others},
  journal={arXiv preprint arXiv:2512.23236},
  year={2025}
}

@article{du2026kernel,
  title={Kernel-Smith: A Unified Recipe for Evolutionary Kernel Optimization},
  author={Du, He and Ge, Qiming and Hu, Jiakai and Yang, Aijun and Cai, Zheng and Huang, Zixian and Yuan, Sheng and Cheng, Qinxiu and Xie, Xinchen and Chen, Yicheng and others},
  journal={arXiv preprint arXiv:2603.28342},
  year={2026}
}

@article{wang2025thetaevolve,
  title={Thetaevolve: Test-time learning on open problems},
  author={Wang, Yiping and Su, Shao-Rong and Zeng, Zhiyuan and Xu, Eva and Ren, Liliang and Yang, Xinyu and Huang, Zeyi and He, Xuehai and Ma, Luyao and Peng, Baolin and others},
  journal={arXiv preprint arXiv:2511.23473},
  year={2025}
}

@article{yuksekgonul2026learning,
  title={Learning to discover at test time},
  author={Yuksekgonul, Mert and Koceja, Daniel and Li, Xinhao and Bianchi, Federico and McCaleb, Jed and Wang, Xiaolong and Kautz, Jan and Choi, Yejin and Zou, James and Guestrin, Carlos and others},
  journal={arXiv preprint arXiv:2601.16175},
  year={2026}
}

@article{ouyang2025kernelbench,
  title={Kernelbench: Can llms write efficient gpu kernels?},
  author={Ouyang, Anne and Guo, Simon and Arora, Simran and Zhang, Alex L and Hu, William and R{\'e}, Christopher and Mirhoseini, Azalia},
  journal={arXiv preprint arXiv:2502.10517},
  year={2025}
}

@article{lin2025can,
  title={Can Language Models Discover Scaling Laws?},
  author={Lin, Haowei and Ye, Haotian and Feng, Wenzheng and Huang, Quzhe and Li, Yujun and Lim, Hubert and Li, Zhengrui and Wang, Xiangyu and Ma, Jianzhu and Liang, Yitao and others},
  journal={arXiv preprint arXiv:2507.21184},
  year={2025}
}

@article{imajuku2025ale,
  title={Ale-bench: A benchmark for long-horizon objective-driven algorithm engineering},
  author={Imajuku, Yuki and Horie, Kohki and Iwata, Yoichi and Aoki, Kensho and Takahashi, Naohiro and Akiba, Takuya},
  journal={arXiv preprint arXiv:2506.09050},
  year={2025}
}

@article{mang2025frontiercs,
  title={FrontierCS: Evolving Challenges for Evolving Intelligence},
  author={Mang, Qiuyang and Chai, Wenhao and Li, Zhifei and Mao, Huanzhi and Zhou, Shang and Du, Alexander and Li, Hanchen and Liu, Shu and Chen, Edwin and Wang, Yichuan and others},
  journal={arXiv preprint arXiv:2512.15699},
  year={2025}
}

@article{shojaee2025llm,
  title={Llm-srbench: A new benchmark for scientific equation discovery with large language models},
  author={Shojaee, Parshin and Nguyen, Ngoc-Hieu and Meidani, Kazem and Farimani, Amir Barati and Doan, Khoa D and Reddy, Chandan K},
  journal={arXiv preprint arXiv:2504.10415},
  year={2025}
}

@incollection{darwin2025origin,
  title={On the origin of species},
  author={Darwin, Charles},
  booktitle={Scientific Methodology in Nineteenth Century Britain},
  pages={133--181},
  year={2025},
  publisher={Routledge}
}

@article{grant2002unpredictable,
  title={Unpredictable evolution in a 30-year study of Darwin's finches},
  author={Grant, Peter R and Grant, B Rosemary},
  journal={science},
  volume={296},
  number={5568},
  pages={707--711},
  year={2002},
  publisher={American Association for the Advancement of Science}
}

@article{shetty2025gso,
  title={Gso: Challenging software optimization tasks for evaluating swe-agents},
  author={Shetty, Manish and Jain, Naman and Liu, Jinjian and Kethanaboyina, Vijay and Sen, Koushik and Stoica, Ion},
  journal={arXiv preprint arXiv:2505.23671},
  year={2025}
}

@misc{gpumode,
  title        = {{GPU} {MODE}},
  author       = {{GPU MODE}},
  howpublished = {\url{https://www.gpumode.com/home}},
  note         = {Accessed: 2026-05-03},
  year         = {2026}
}

@misc{erdosproblems,
  title        = {Erdos Problems},
  author       = {Bloom, Thomas},
  howpublished = {\url{https://www.erdosproblems.com/}},
  note         = {Accessed: 2026-05-03},
  year         = {2026}
}

@article{feng2026semi,
  title={Semi-Autonomous Mathematics Discovery with Gemini: A Case Study on the Erd$\backslash$H $\{$o$\}$ s Problems},
  author={Feng, Tony and Trinh, Trieu and Bingham, Garrett and Kang, Jiwon and Zhang, Shengtong and Kim, Sang-hyun and Barreto, Kevin and Schildkraut, Carl and Jung, Junehyuk and Seo, Jaehyeon and others},
  journal={arXiv preprint arXiv:2601.22401},
  year={2026}
}

@article{luecken2025defining,
  title={Defining and benchmarking open problems in single-cell analysis},
  author={Luecken, Malte D and Gigante, Scott and Burkhardt, Daniel B and Cannoodt, Robrecht and Strobl, Daniel C and Markov, Nikolay S and Zappia, Luke and Palla, Giovanni and Lewis, Wesley and Dimitrov, Daniel and others},
  journal={Nature Biotechnology},
  volume={43},
  number={7},
  pages={1035--1040},
  year={2025},
  publisher={Nature Publishing Group US New York}
}

@inproceedings{longpre2023flan,
  title={The flan collection: Designing data and methods for effective instruction tuning},
  author={Longpre, Shayne and Hou, Le and Vu, Tu and Webson, Albert and Chung, Hyung Won and Tay, Yi and Zhou, Denny and Le, Quoc V and Zoph, Barret and Wei, Jason and others},
  booktitle={International conference on machine learning},
  pages={22631--22648},
  year={2023},
  organization={PMLR}
}

@misc{qwen35blog,
    title = {Qwen3.5: Accelerating Productivity with Native Multimodal Agents},
    url = {https://qwen.ai/blog?id=qwen3.5},
    author = {Qwen Team},
    month = {February},
    year = {2026}
}

@article{ye2026evaluation,
  title={Evaluation-driven Scaling for Scientific Discovery},
  author={Ye, Haotian and Lin, Haowei and Tang, Jingyi and Luo, Yizhen and Yang, Caiyin and Su, Chang and Thapa, Rahul and Yang, Rui and Liu, Ruihua and Li, Zeyu and others},
  journal={arXiv preprint arXiv:2604.19341},
  year={2026}
}

@inproceedings{zheng2024llamafactory,
  title={Llamafactory: Unified efficient fine-tuning of 100+ language models},
  author={Zheng, Yaowei and Zhang, Richong and Zhang, Junhao and Ye, Yanhan and Luo, Zheyan},
  booktitle={Proceedings of the 62nd annual meeting of the association for computational linguistics (volume 3: system demonstrations)},
  pages={400--410},
  year={2024}
}

@article{press2025algotune,
  title={AlgoTune: Can Language Models Speed Up General-Purpose Numerical Programs?},
  author={Press, Ori and Amos, Brandon and Zhao, Haoyu and Wu, Yikai and Ainsworth, Samuel K and Krupke, Dominik and Kidger, Patrick and Sajed, Touqir and Stellato, Bartolomeo and Park, Jisun and others},
  journal={arXiv preprint arXiv:2507.15887},
  year={2025}
}

@book{vygotsky2011interaction,
  title={Interaction between learning and development},
  author={Vygotsky, Lev and others},
  year={2011},
  publisher={Link{\"o}pings universitet Link{\"o}ping, Sweden}
}

@misc{autolab-2026,
  title   = {AutoLab: Can Models Begin to Participate
             in the Loops That Drive Scientific
             and Engineering Progress?},
  author  = {AutoLab Team},
  year    = {2026},
  url     = {https://github.com/autolabhq/autolab}
}

@article{erdHos1955some,
  title={Some remarks on number theory},
  author={Erd{\H{o}}s, Paul},
  journal={Riveon Lematematika},
  volume={9},
  pages={45--48},
  year={1955}
}

@article{white2023new,
  title={A new bound for Erd{\H{o}}s’ minimum overlap problem},
  author={White, Ethan Patrick},
  journal={Acta Arithmetica},
  volume={208},
  pages={235--255},
  year={2023},
  publisher={Instytut Matematyczny Polskiej Akademii Nauk}
}

@article{kirkpatrick1983optimization,
  title={Optimization by simulated annealing},
  author={Kirkpatrick, Scott and Gelatt Jr, C Daniel and Vecchi, Mario P},
  journal={science},
  volume={220},
  number={4598},
  pages={671--680},
  year={1983},
  publisher={American association for the advancement of science}
}

@inproceedings{lee2024collavo,
  title={Collavo: Crayon large language and vision model},
  author={Lee, Byung-Kwan and Park, Beomchan and Kim, Chae Won and Ro, Yong Man},
  booktitle={Findings of the Association for Computational Linguistics: ACL 2024},
  pages={1121--1138},
  year={2024}
}

@inproceedings{lee2024moai,
  title={Moai: Mixture of all intelligence for large language and vision models},
  author={Lee, Byung-Kwan and Park, Beomchan and Won Kim, Chae and Man Ro, Yong},
  booktitle={European Conference on Computer Vision},
  pages={273--302},
  year={2024},
  organization={Springer}
}

@article{lee2024meteor,
  title={Meteor: Mamba-based traversal of rationale for large language and vision models},
  author={Lee, Byung-Kwan and Kim, Chae Won and Park, Beomchan and Ro, Yong Man},
  journal={Advances in Neural Information Processing Systems},
  volume={37},
  pages={40278--40315},
  year={2024}
}

@inproceedings{lee2025vlsi,
  title={Vlsi: Verbalized layers-to-interactions from large to small vision language models},
  author={Lee, Byung-Kwan and Hachiuma, Ryo and Wang, Yu-Chiang Frank and Ro, Yong Man and Wu, Yueh-Hua},
  booktitle={Proceedings of the IEEE/CVF Conference on Computer Vision and Pattern Recognition},
  pages={29545--29557},
  year={2025}
}

@article{lee2026unified,
  title={Unified reinforcement and imitation learning for vision-language models},
  author={Lee, Byung-Kwan and Hachiuma, Ryo and Ro, Yong Man and Wang, Frank and Wu, Yueh-Hua},
  journal={Advances in Neural Information Processing Systems},
  volume={38},
  pages={156508--156534},
  year={2026}
}

@article{kang2026agent,
  title={Agent Explorative Policy Optimization for Multimodal Agentic Reasoning},
  author={Kang, Minki and Diao, Shizhe and Hachiuma, Ryo and Hwang, Sung Ju and Molchanov, Pavlo and Wang, Yu-Chiang Frank and Lee, Byung-Kwan},
  journal={arXiv preprint arXiv:2605.28774},
  year={2026}
}

@inproceedings{lee2026masking,
  title={Masking teacher and reinforcing student for distilling vision-language models},
  author={Lee, Byung-Kwan and Wang, Yu-Chiang Frank and Hachiuma, Ryo},
  booktitle={Proceedings of the IEEE/CVF Conference on Computer Vision and Pattern Recognition},
  pages={10126--10141},
  year={2026}
}

@article{lee2024phantom,
  title={Phantom of latent for large language and vision models},
  author={Lee, Byung-Kwan and Chung, Sangyun and Kim, Chae Won and Park, Beomchan and Ro, Yong Man},
  journal={arXiv preprint arXiv:2409.14713},
  year={2024}
}

@article{lee2025genrecal,
  title={Genrecal: Generation after recalibration from large to small vision-language models},
  author={Lee, Byung-Kwan and Hachiuma, Ryo and Ro, Yong Man and Wang, Yu-Chiang Frank and Wu, Yueh-Hua},
  journal={arXiv preprint arXiv:2506.15681},
  year={2025}
}

@article{lee2024trol,
  title={Trol: Traversal of layers for large language and vision models},
  author={Lee, Byung-Kwan and Chung, Sangyun and Kim, Chae Won and Park, Beomchan and Ro, Yong Man},
  journal={arXiv preprint arXiv:2406.12246},
  year={2024}
}

@inproceedings{lee2025multiverse,
  title={Multiverse: A multi-turn conversation benchmark for evaluating large vision and language models},
  author={Lee, Young-Jun and Lee, Byung-Kwan and Zhang, Jianshu and Hwang, Yechan and Ko, Byungsoo and Kim, Han-Gyu and Yao, Dongyu and Rong, Xuankun and Joo, Eojin and Han, Seung-Ho and others},
  booktitle={Proceedings of the IEEE/CVF International Conference on Computer Vision},
  pages={708--719},
  year={2025}
}

@article{lee2025refinebench,
  title={RefineBench: Evaluating Refinement Capability of Language Models via Checklists},
  author={Lee, Young-Jun and Kim, Seungone and Lee, Byung-Kwan and Moon, Minkyeong and Hwang, Yechan and Kim, Jong Myoung and Neubig, Graham and Welleck, Sean and Choi, Ho-Jin},
  journal={arXiv preprint arXiv:2511.22173},
  year={2025}
}

@inproceedings{lee2026recursive,
  title={Recursive think-answer process for llms and vlms},
  author={Lee, Byung-Kwan and Chee, Youngchae and Ro, Yong Man},
  booktitle={Proceedings of the IEEE/CVF Conference on Computer Vision and Pattern Recognition},
  pages={9608--9621},
  year={2026}
}

@article{cho2026spatialclaw,
  title={SpatialClaw: Rethinking Action Interface for Agentic Spatial Reasoning},
  author={Cho, Seokju and Hachiuma, Ryo and Badki, Abhishek and Su, Hang and Lee, Byung-Kwan and Song, Chan Hee and Liu, Sifei and Radhakrishnan, Subhashree and Kim, Seungryong and Wang, Yu-Chiang Frank and others},
  journal={arXiv preprint arXiv:2606.13673},
  year={2026}
}

@article{yu2026hide,
  title={Hide to See: Reasoning-prefix Masking for Visual-anchored Thinking in VLM Distillation},
  author={Yu, Seonghoon and Nam, Dongjun and Lee, Byung-Kwan and Son, Jeany},
  journal={arXiv preprint arXiv:2605.11651},
  year={2026}
}

@article{kim2026and,
  title={Why and when visual token pruning fails? a study on relevant visual information shift in mllms decoding},
  author={Kim, Jiwan and Kim, Kibum and Kim, Wonjoong and Lee, Byung-Kwan and Park, Chanyoung},
  journal={arXiv preprint arXiv:2604.12358},
  year={2026}
}

@article{ma2026orion,
  title={Orion: Towards Lab Automation with Computer-Using Agents},
  author={Ma, Chang and Trinh, Linh and Bucci, Matt and Regev, Aviv and Wang, Hanchen},
  journal={bioRxiv},
  pages={2026--06},
  year={2026},
  publisher={Cold Spring Harbor Laboratory}
}

@article{sun2025scienceboard,
  title={Scienceboard: Evaluating multimodal autonomous agents in realistic scientific workflows},
  author={Sun, Qiushi and Liu, Zhoumianze and Ma, Chang and Ding, Zichen and Xu, Fangzhi and Yin, Zhangyue and Zhao, Haiteng and Wu, Zhenyu and Cheng, Kanzhi and Liu, Zhaoyang and others},
  journal={arXiv preprint arXiv:2505.19897},
  year={2025}
}

@inproceedings{lee2024dialogcc,
  title={Dialogcc: An automated pipeline for creating high-quality multi-modal dialogue dataset},
  author={Lee, Young-Jun and Ko, Byungsoo and Kim, Han-Gyu and Hyeon, Jonghwan and Choi, Ho-Jin},
  booktitle={Proceedings of the 2024 Conference of the North American Chapter of the Association for Computational Linguistics: Human Language Technologies (Volume 1: Long Papers)},
  pages={1938--1963},
  year={2024}
}

@inproceedings{lee2024stark,
  title={Stark: Social long-term multi-modal conversation with persona commonsense knowledge},
  author={Lee, Young-Jun and Lee, Dokyong and Youn, Junyoung and Oh, Kyeong-Jin and Ko, Byungsoo and Hyeon, Jonghwan and Choi, Ho-Jin},
  booktitle={Findings of the Association for Computational Linguistics: EMNLP 2024},
  pages={12137--12162},
  year={2024}
}

@article{lee2024thanos,
  title={Thanos: Enhancing conversational agents with skill-of-mind-infused large language model},
  author={Lee, Young-Jun and Lee, Dokyong and Youn, Junyoung and Oh, Kyeongjin and Choi, Ho-Jin},
  journal={arXiv preprint arXiv:2411.04496},
  year={2024}
}

@inproceedings{lee2024large,
  title={Large Language Models can Share Images, Too!},
  author={Lee, Young-Jun and Lee, Dokyong and Sung, Joo-won and Hyeon, Jonghwan and Choi, Ho-Jin},
  booktitle={Findings of the Association for Computational Linguistics: ACL 2024},
  pages={692--713},
  year={2024}
}
}

\clearpage


\appendix

\section{Broader Impacts} \label{supp_sec:broader_impacts}

EFT democratizes LLM-driven discovery by transferring optimization capabilities from expensive proprietary models to small open-weight models, reducing search costs, enabling fully local discovery pipelines, and providing \datasetName as a reusable resource for future research. At the same time, these capabilities may be misused for harmful optimization objectives, reward hacking, or over-reliance on automatically generated discoveries without sufficient verification. To mitigate these risks, we train only on public scientific and engineering benchmarks, preserve the upstream safety properties of base models, and recommend human oversight, scoring-function auditing, and trajectory logging for downstream deployment.

\section{Further Details and Discussion} \label{supp_sec:discussion}

\subsection{Why is the model named \modelName?} \label{supp_sec:model_naming}

Darwin’s finches, despite belonging to the same species, have evolved differently in response to the diverse ecological environments of the Galápagos Islands~\citep{darwin2025origin,grant2002unpredictable}. This observation illustrates their remarkable ability to adapt to a wide range of environmental conditions and to thrive within them.

Inspired by this phenomenon, we name our model \modelName. Analogous to Darwin’s finches, our model is designed to adapt across diverse environments—here corresponding to different optimization tasks—and to effectively operate within them. In particular, it reflects the model’s ability to flexibly adapt to various tasks (\eg mathematics, kernel engineering, biology denoising) and to discover better solutions within each task setting.

\subsection{Definition of Scaffolds} \label{supp_sec:scaffold_definition}

The term \textit{scaffold} is borrowed from developmental psychology. Vygotsky~\citep{vygotsky2011interaction} characterizes scaffolding as external support that elevates a learner's \textit{assisted performance} beyond their unaided reach, and further distinguishes between progress driven by \textit{externally supplied} structure and progress driven by the learner's \textit{internalization} of past experience. Following this distinction, we taxonomize evolutionary search scaffolds into \textbf{supervised} and \textbf{unsupervised} scaffolds.

\paragraph{Definition: Supervised Scaffold.}
A supervised scaffold (\ie, search-based)~\citep{openevolve,lange2025shinkaevolve,cemri2026adaevolve,liu2026evox,skydiscover2026} keeps the mutation operator's parameters $\theta$ fixed and drives discovery through a hand-engineered modular framework that compensates for the LLM's limited intrinsic capacity for discovery. It plays the role of an external tutor, as the policies governing how candidate solutions are selected and stored are typically designed by human experts.

\paragraph{Definition: Unsupervised Scaffold.}
An unsupervised scaffold (\ie, learning-based)~\citep{wang2025thetaevolve,yuksekgonul2026learning} updates $\theta$ at test time via reinforcement learning (RL) over the LLM's self-generated experiences. Prior works still retain a database and a selection strategy (\eg, PUCT), but the surrounding framework is comparatively lightweight, corresponding to Vygotsky's \textit{internalization} phase.

Strong performance under a supervised scaffold is not evidence that $\mathcal{M}_{\theta}$ has \textit{internalized} discovery: since $\theta$ is frozen, any gain is jointly attributable to the LLM and its external framework, and corresponds to Vygotsky's assisted performance. Genuine internalization requires improvements to accrue to $\mathcal{M}_{\theta}$ itself---persisting even after the scaffolding is stripped away---which motivates the unsupervised regime. In this work, our goal is to equip the LLM with internalized discovery capability, enabling it to operate effectively under both supervised and unsupervised scaffolds.

\begin{table}[!t]
\centering
\caption{Key differences across three standard tasks: Reasoning vs. Agentic vs. Optimization.}
\label{supp_tab:task_difference}
\vspace{5pt}
\renewcommand{\arraystretch}{1.3}
\small

\begin{adjustbox}{width=\linewidth}
\newcolumntype{C}[1]{>{\centering\arraybackslash}p{#1}}
\begin{tabular}{@{}l C{4.2cm} C{4.2cm} C{4.2cm}@{}}

\toprule
\textbf{Characteristic} & \textbf{Reasoning} & \textbf{Agentic} & \textbf{Optimization} \\
\midrule

\multicolumn{4}{l}{\cellcolor{gray!10}\textit{Problem}} \\

Objective
  & Solving a specific, well-defined problem
  & Complete real-world task
  & Discovering novel solutions, algorithms, or structures \\

Ground Truth
  & Known (Solution)
  & Known (Acceptance Criteria)
  & Unknown \\

Structure
  & Closed-form problem solving
  & Constrained execution
  & Open-ended problem solving \\

Complexity
  & P or NP (decidable)
  & Compositional
  & NP-Hard \\

Success Criterion
  & Solve --- match ground truth
  & Complete --- satisfy acceptance criteria
  & Improve --- exceed previous best-known \\

Expertise
  & Expert-curated exam problems designed with known answers
  & Real-world application
  & Expert-verified open problems \\

\midrule
\multicolumn{4}{l}{\cellcolor{gray!10}\textit{Evaluation}} \\

Verification
  & Deterministic, Binary
  & Deterministic, Binary
  & Deterministic, Continuous \\

Metric
  & Accuracy
  & Resolve rate
  & Relative improvement \\

\midrule
\multicolumn{4}{l}{\cellcolor{gray!10}\textit{Execution}} \\

Env.\ Interaction
  & \xmark
  & \cmark
  & \cmark \\

Time Horizon
  & Short
  & Mid
  & Long \\

Role of LLM
  & Reasoning Engine
  & Planner and Executor
  & Mutation Operator \\

Search Space
  & Discrete answer space
  & Action trajectories over system state
  & Combinatorial / continuous program space \\

\bottomrule
\end{tabular}
\end{adjustbox}

\end{table}

\subsection{Distinguishing Optimization from Reasoning and Agentic Tasks} \label{supp_sec:task_comparison}

We position optimization (or discovery) tasks as fundamentally distinct from both reasoning and agentic tasks, as summarized in Table~\ref{supp_tab:task_difference}. 

Unlike reasoning tasks, which assume well-defined problems with known ground-truth solutions, optimization tasks are inherently open-ended: the objective is not to recover a known answer, but to discover improved or entirely novel solutions. This lack of ground truth shifts the success criterion from correctness to relative improvement over prior best-known results. 

Compared to agentic tasks, which emphasize executing sequences of actions to satisfy predefined acceptance criteria in real-world environments, optimization tasks require searching over a substantially larger and more abstract solution space (e.g., programs, algorithms, or mathematical constructions). As a result, they exhibit longer time horizons and demand iterative refinement rather than single-pass execution.

These differences lead to a distinct evaluation paradigm. While reasoning and agentic tasks are typically evaluated using binary success signals, optimization tasks rely on deterministic but continuous metrics that enable partial progress to be measured and accumulated over time.

Taken together, optimization tasks necessitate a different role for language models: instead of acting purely as a reasoning engine or a planner, the model serves as a mutation operator within an evolutionary search process, iteratively proposing candidate solutions that can be evaluated and improved. This distinction motivates the design of \modelName, which is specifically tailored to support discovery-driven optimization.

\section{Additional Details of Dataset Construction Method} \label{supp_sec:additional_dataset_construction}

\subsection{Systematic Error Breakdown Analysis}

\begin{table}[ht!]
\centering
\setlength{\tabcolsep}{4pt}
\renewcommand{\arraystretch}{1.15}
\caption{Breakdown of system-level errors filtered out (6,321 trajectories). Counts and percentages are calculated over the total of 6,321 trajectories.}
\label{supp_tab:system_error_breakdown}
\footnotesize
\begin{adjustbox}{max width=\textwidth}
\begin{tabular}{@{}p{4.6cm} r r p{8.0cm}@{}}
\toprule
\textbf{Error category} & \textbf{Count} & \textbf{\%} & \textbf{Description} \\
\midrule
\texttt{parent\_missing\_combined\_score}
& 3{,}641 & 57.60
& Parent program itself never produced a \texttt{combined\_score}, so the parent$\to$child improvement delta is undefined regardless of what the child did. \\
\texttt{artifact\_error\_type:timeout}
& \phantom{0,}901 & 14.25
& Evaluator wall-clock timeout: child program ran past its time budget (\texttt{artifacts.timeout = True}, \texttt{error\_type = "timeout"}). \\
\texttt{failure\_stage:correctness}
& \phantom{0,}794 & 12.56
& Child program ran to completion but produced numerically incorrect outputs (the evaluator's correctness check rejected the result). \\
\texttt{child\_metrics\_error\_string}
& \phantom{0,}605 & \phantom{0}9.57
& \texttt{child\_metrics.error} holds a human-readable failure message (e.g.\ \texttt{"C1 mismatch: reported X, computed Y"}) rather than a clean numeric metric. \\
\texttt{artifact\_error\_type:TimeoutError}
& \phantom{0,}164 & \phantom{0}2.59
& Async \texttt{TimeoutError} raised from the evaluator's task wrapper (typically during \texttt{cascade\_setup}, \texttt{stage1}, or \texttt{stage2}). \\
\texttt{failure\_stage:benchmark}
& \phantom{0,}147 & \phantom{0}2.33
& Failure inside the benchmark/timing harness \emph{after} the correctness check passed (e.g.\ a crash during the performance-measurement loop). \\
\texttt{artifact\_error\_string}
& \phantom{00,}61 & \phantom{0}0.97
& \texttt{artifacts.error} contains a runtime exception string + traceback (e.g.\ \texttt{NameError}, syntax error, JAX trace error) with no structured \texttt{error\_type} or \texttt{failure\_stage}. \\
\texttt{failure\_stage:syntax\_check}
& \phantom{000,}5 & \phantom{0}0.08
& Generated code failed the pre-execution syntax check (could not be parsed or compiled). \\
\texttt{failure\_stage:import}
& \phantom{000,}2 & \phantom{0}0.03
& Generated module failed to import (missing symbol, \texttt{ImportError}, or top-level exception during import). \\
\texttt{failure\_stage:validation}
& \phantom{000,}1 & \phantom{0}0.02
& Generated solution failed schema/shape validation before execution. \\
\midrule
\textbf{Total} & \textbf{6{,}321} & \textbf{100.00} & \\
\bottomrule
\end{tabular}
\end{adjustbox}
\end{table}

\subsection{Benchmark Licenses}
\label{supp_sec:benchmark_license}

We construct our seed tasks from publicly available benchmarks, all of which are released under permissive open-source licenses (\eg MIT, Apache 2.0, or CC-BY 4.0). Table~\ref{supp_tab:benchmark_license} summarizes the license terms for the corresponding GitHub repositories and associated datasets for each benchmark. Our use of these resources is limited to non-commercial academic research and complies with the terms of their respective licenses, including all attribution requirements. Accordingly, \datasetName is released under the CC-BY 4.0 license, while our code and \modelName weights are released under the Apache 2.0 license. Both are compatible with the licenses of all upstream benchmarks listed above. In particular, the tasks sourced from ALE-Bench~\citep{imajuku2025ale} are derived from a dataset released under the CC BY-ND 4.0 license. We sincerely thank the AtCoder team for granting permission to publicly release the evolutionary search trajectories derived from their CC BY-ND 4.0 licensed dataset.

\begin{table}[h]
\centering
\caption{Benchmark Licenses}
\label{supp_tab:benchmark_license}
\vspace{5pt}
\begin{tabular}{lll}
\toprule
\textbf{Benchmark} & \textbf{Github} & \textbf{Dataset} \\
\midrule
LLM-SRBench~\citep{shojaee2025llm} & MIT License & - \\
FrontierCS~\citep{mang2025frontiercs} & MIT License & Apache 2.0 \\
ALE-Bench~\citep{imajuku2025ale} & Apache 2.0 & CC BY-ND 4.0 \\
GPU Mode & - & - \\
AlgoTune~\citep{press2025algotune} & MIT License & MIT License \\
AlphaEvolve's Math Problems~\citep{novikov2025alphaevolve} & Apache License 2.0 & CC-BY 4.0 \\
Function Minimization~\citep{openevolve} & Apache License 2.0 & Apache License 2.0 \\
\bottomrule
\end{tabular}
\end{table}

\section{Additional Implementation Details} \label{supp_sec:additional_implementation_details}

\paragraph{Implementation details for \modelName-8B nanodiscover runs.}

nanodiscover is an open-source reproduction of TTT-Discover that does not depend on the Tinker API. It is publicly available at \url{https://github.com/cheongalc/nanodiscover}. Each search epoch in nanodiscover consists of five stages mirroring the TTT-Discover pipeline: (1) sampling parent solutions from the archive, (2) generating child solutions from parent solutions, (3) evaluating child solutions, (4) updating the archive, and (5) test-time training. Ray Data LLM (which orchestrates vLLM under the hood) is used for step (2), while DeepSpeed is used for step (5). 
Unless otherwise noted, all hyperparameters were matched to TTT-Discover as closely as possible.

All runs were conducted for 50 epochs on a single node with 4 GPUs and 96 logical CPU cores. For the Erd\H{o}s task, the prompt informed the model that the evaluation budget was 1000 seconds, while the actual timeout was set to 1100 seconds. For both circle-packing tasks, the model was not informed of the evaluation budget, and the actual timeout was 530 seconds. These timing configurations follow the TTT-Discover setup.

\newpage

\section{Full List of Optimization Tasks in \datasetName} \label{supp_sec:full_task_list}

{\setlength{\tabcolsep}{4pt}%
\renewcommand{\arraystretch}{1.15}%
\footnotesize
\begin{longtable}{@{}p{4.0cm} p{2.1cm} p{8.0cm}@{}}
\caption{%
Per-task descriptions for the \textbf{Mathematical Discovery} task group (14 tasks) evaluated under \datasetName{}.%
}
\label{tab:task-desc-math} \\
\toprule
\textbf{Task Name} & \textbf{Objective} & \textbf{Description} \\
\midrule
\endfirsthead

\multicolumn{3}{@{}l}{\textit{Table~\ref{tab:task-desc-math} (continued)}} \\
\toprule
\textbf{Task Name} & \textbf{Objective} & \textbf{Description} \\
\midrule
\endhead

\midrule
\multicolumn{3}{r@{}}{\textit{Continued on next page}} \\
\endfoot

\bottomrule
\endlastfoot

\texttt{circle\_\allowbreak{}packing\_\allowbreak{}rect}
& $\max\sum r_i$
& Pack equal-radius circles inside an axis-aligned rectangle without overlap; maximize the common radius under boundary and non-overlap constraints. \\

\texttt{erdos\_\allowbreak{}min\_\allowbreak{}overlap}
& $\min M(n)$
& Construct a witness function (step function on $[-1,1]$) that gives a constructive upper bound on Erd\H{o}s's minimum-overlap constant $M(n)$. \\

\texttt{first\_\allowbreak{}autocorr\_\allowbreak{}ineq}
& $\min C_1$
& First autocorrelation inequality: minimize $\lVert f * f\rVert_\infty$ over functions $f\ge 0$ supported on $[-1/4,1/4]$ with $\int f = 1$. \\

\texttt{second\_\allowbreak{}autocorr\_\allowbreak{}ineq}
& $\max C_2$
& Second autocorrelation inequality: extremal $\lVert f*f\rVert$ under prescribed support and mass constraints. \\

\texttt{third\_\allowbreak{}autocorr\_\allowbreak{}ineq}
& $\min C_3$
& Third autocorrelation inequality: tighten the upper bound on the third autocorrelation constant arising in additive combinatorics. \\

\texttt{heilbronn\_\allowbreak{}triangle}
& $\max\min$ area
& Place $n$ points in the unit square $[0,1]^2$; maximize the area of the smallest triangle formed by any three points. \\

\texttt{heilbronn\_\allowbreak{}convex\_\allowbreak{}13}
& $\max\min$ area
& Heilbronn-on-a-convex-region variant with $n{=}13$ points: maximize the minimum convex-hull area over all subsets of $k>3$ points. \\

\texttt{heilbronn\_\allowbreak{}convex\_\allowbreak{}14}
& $\max\min$ area
& Heilbronn-on-a-convex-region variant with $n{=}14$ points (otherwise as above). \\

\texttt{hexagon\_\allowbreak{}packing\_\allowbreak{}11}
& $\min$ enclosing
& Pack $n{=}11$ unit regular hexagons inside the smallest enclosing regular hexagon. \\

\texttt{hexagon\_\allowbreak{}packing\_\allowbreak{}12}
& $\min$ enclosing
& Pack $n{=}12$ unit regular hexagons inside the smallest enclosing regular hexagon. \\

\texttt{minimizing\_\allowbreak{}max\_\allowbreak{}min\_\allowbreak{}dist\_\allowbreak{}2}
& $\max\, d_{\min}/d_{\max}$
& Place points in $[0,1]^2$ so that the ratio of minimum to maximum pairwise distance is as close to 1 as possible (uniform 2D point distribution). \\

\texttt{minimizing\_\allowbreak{}max\_\allowbreak{}min\_\allowbreak{}dist\_\allowbreak{}3}
& $\max\, d_{\min}/d_{\max}$
& Same as above but for points in $[0,1]^3$ (uniform 3D point distribution). \\

\texttt{signal\_\allowbreak{}processing}
& multi-objective
& Design a causal real-time filter for a noisy non-stationary time series, balancing fidelity, smoothness, lag, and false-trend detection. \\

\texttt{sums\_\allowbreak{}diffs\_\allowbreak{}finite\_\allowbreak{}sets}
& $\min |A{+}A|/|A{-}A|$
& Construct a finite set $A\subset\mathbb{Z}$ minimizing the ratio between sumset and difference-set cardinalities (additive combinatorics). \\

\end{longtable}
}%

\bigskip

{\setlength{\tabcolsep}{4pt}%
\renewcommand{\arraystretch}{1.15}%
\footnotesize
\begin{longtable}{@{}p{4.0cm} p{2.1cm} p{8.0cm}@{}}
\caption{%
Per-task descriptions for the \textbf{Numerical Algorithm Optimization} task group (46 tasks) evaluated under \datasetName{}.%
}
\label{tab:task-desc-algotune} \\
\toprule
\textbf{Task Name} & \textbf{Objective} & \textbf{Description} \\
\midrule
\endfirsthead

\multicolumn{3}{@{}l}{\textit{Table~\ref{tab:task-desc-algotune} (continued)}} \\
\toprule
\textbf{Task Name} & \textbf{Objective} & \textbf{Description} \\
\midrule
\endhead

\midrule
\multicolumn{3}{r@{}}{\textit{Continued on next page}} \\
\endfoot

\bottomrule
\endlastfoot

\texttt{aes\_\allowbreak{}gcm\_\allowbreak{}encryption}
& max speedup
& Speed up AES-GCM authenticated encryption against a \texttt{cryptography}-library reference. \\

\texttt{affine\_\allowbreak{}transform\_\allowbreak{}2d}
& max speedup
& Speed up 2D affine image warping with cubic-spline interpolation against a SciPy reference. \\

\texttt{aircraft\_\allowbreak{}wing\_\allowbreak{}design}
& max speedup
& Speed up the geometric-programming aircraft-wing design problem (drag minimization under aerodynamic constraints). \\

\texttt{articulation\_\allowbreak{}points}
& max speedup
& Find all articulation points in an undirected graph (cut vertices whose removal disconnects the graph). \\

\texttt{base64\_\allowbreak{}encoding}
& max speedup
& Speed up Base64 encoding of binary data against a \texttt{base64} stdlib reference. \\

\texttt{battery\_\allowbreak{}scheduling}
& max speedup
& Convex battery charge/discharge scheduling under price, capacity, and ramp constraints (CVXPY reference). \\

\texttt{chacha\_\allowbreak{}encryption}
& max speedup
& Speed up ChaCha20 stream-cipher encryption against a \texttt{cryptography}-library reference. \\

\texttt{channel\_\allowbreak{}capacity}
& max speedup
& Compute the channel capacity of a discrete memoryless channel by maximizing mutual information (CVXPY reference). \\

\texttt{chebyshev\_\allowbreak{}center}
& max speedup
& Find the center of the largest inscribed ball of a polyhedron $P=\{x\mid a_i^\top x\le b_i\}$ via an LP. \\

\texttt{cholesky\_\allowbreak{}factorization}
& max speedup
& Speed up dense Cholesky factorization of a symmetric positive-definite matrix. \\

\texttt{clustering\_\allowbreak{}outliers}
& max speedup
& Outlier-robust clustering of points in multidimensional space (HDBSCAN-style reference). \\

\texttt{communicability}
& max speedup
& Compute communicability $C(u,v)$ between all node pairs in an undirected graph (NetworkX reference). \\

\texttt{convex\_\allowbreak{}hull}
& max speedup
& Compute the convex hull of a 2D point set (smallest convex polygon containing all points). \\

\texttt{convolve2d\_\allowbreak{}full\_\allowbreak{}fill}
& max speedup
& Speed up 2D convolution with full-output, fill boundary against \texttt{scipy.signal.convolve2d}. \\

\texttt{convolve\_\allowbreak{}1d}
& max speedup
& Speed up 1D convolution over a list of array pairs against a SciPy reference. \\

\texttt{correlate\_\allowbreak{}1d}
& max speedup
& Speed up 1D cross-correlation over a list of array pairs against a SciPy reference. \\

\texttt{count\_\allowbreak{}connected\_\allowbreak{}components}
& max speedup
& Count connected components of an undirected edge-list graph. \\

\texttt{count\_\allowbreak{}riemann\_\allowbreak{}zeta\_\allowbreak{}zeros}
& max speedup
& Count nontrivial zeros of the Riemann zeta function with imaginary part in $(0,t]$. \\

\texttt{cumulative\_\allowbreak{}simpson\_\allowbreak{}1d}
& max speedup
& Cumulative integral of a 1D function via Simpson's rule (SciPy reference). \\

\texttt{cumulative\_\allowbreak{}simpson\_\allowbreak{}multid}
& max speedup
& Cumulative integral along the last axis of a multidimensional array via Simpson's rule. \\

\texttt{dct\_\allowbreak{}type\_\allowbreak{}I\_\allowbreak{}scipy\_\allowbreak{}fftpack}
& max speedup
& Speed up Type-I Discrete Cosine Transform against \texttt{scipy.fftpack}. \\

\texttt{delaunay}
& max speedup
& Compute Delaunay triangulation of a 2D point set (SciPy reference). \\

\texttt{dijkstra\_\allowbreak{}from\_\allowbreak{}indices}
& max speedup
& Shortest paths from a subset of source nodes to all others on a weighted undirected CSR graph. \\

\texttt{dst\_\allowbreak{}type\_\allowbreak{}II\_\allowbreak{}scipy\_\allowbreak{}fftpack}
& max speedup
& Speed up Type-II Discrete Sine Transform on a 2D array against \texttt{scipy.fftpack}. \\

\texttt{dynamic\_\allowbreak{}assortment\_\allowbreak{}planning}
& max speedup
& Dynamic assortment planning over $T$ periods with $N$ products under per-product capacities; maximize expected revenue. \\

\texttt{earth\_\allowbreak{}movers\_\allowbreak{}distance}
& max speedup
& Solve the optimal-transport problem between two histograms with a given cost matrix (POT reference). \\

\texttt{edge\_\allowbreak{}expansion}
& max speedup
& Compute the edge expansion $\partial S/|S|$ for a node subset $S$ in a directed graph. \\

\texttt{eigenvalues\_\allowbreak{}complex}
& max speedup
& Eigenvalues of a real square matrix that may have complex eigenvalues (LAPACK reference). \\

\texttt{eigenvalues\_\allowbreak{}real}
& max speedup
& Eigenvalues of a symmetric real matrix (LAPACK reference). \\

\texttt{eigenvectors\_\allowbreak{}complex}
& max speedup
& Eigenpairs of a real square matrix (complex eigenvalues/eigenvectors). \\

\texttt{eigenvectors\_\allowbreak{}real}
& max speedup
& Eigenpairs of a real symmetric matrix (orthonormal eigenvectors). \\

\texttt{elementwise\_\allowbreak{}integration}
& max speedup
& Elementwise definite integration of Wright's Bessel function across an array of arguments. \\

\texttt{fft\_\allowbreak{}cmplx\_\allowbreak{}scipy\_\allowbreak{}fftpack}
& max speedup
& $N$-dimensional complex FFT of a complex matrix against \texttt{scipy.fftpack}. \\

\texttt{fft\_\allowbreak{}convolution}
& max speedup
& FFT-based convolution of two signals (overlap-add / circular-FFT reference). \\

\texttt{fft\_\allowbreak{}real\_\allowbreak{}scipy\_\allowbreak{}fftpack}
& max speedup
& $N$-dimensional FFT of a real-valued matrix against \texttt{scipy.fftpack}. \\

\texttt{firls}
& max speedup
& Design a least-squares FIR filter for given frequency-band edges (SciPy \texttt{firls} reference). \\

\texttt{generalized\_\allowbreak{}eigenvalues\_\allowbreak{}complex}
& max speedup
& Generalized eigenvalues of $(A,B)$ where $A,B$ are real and the spectrum may be complex. \\

\texttt{generalized\_\allowbreak{}eigenvalues\_\allowbreak{}real}
& max speedup
& Generalized eigenvalues of $(A,B)$ with real spectrum (symmetric/positive-definite case). \\

\texttt{generalized\_\allowbreak{}eigenvectors\_\allowbreak{}complex}
& max speedup
& Generalized eigenpairs of $(A,B)$ with complex spectrum. \\

\texttt{generalized\_\allowbreak{}eigenvectors\_\allowbreak{}real}
& max speedup
& Generalized eigenpairs of $(A,B)$ with real spectrum. \\

\texttt{graph\_\allowbreak{}global\_\allowbreak{}efficiency}
& max speedup
& Average inverse shortest-path length over all node pairs of an undirected graph (NetworkX reference). \\

\texttt{graph\_\allowbreak{}isomorphism}
& max speedup
& Find a node mapping between two isomorphic undirected graphs (NetworkX VF2 reference). \\

\texttt{graph\_\allowbreak{}laplacian}
& max speedup
& Compute the combinatorial or symmetric-normalized Laplacian of a sparse undirected graph. \\

\texttt{group\_\allowbreak{}lasso}
& max speedup
& Logistic regression with group-lasso penalty over $J$ feature groups (CVXPY reference). \\

\texttt{gzip\_\allowbreak{}compression}
& max speedup
& Speed up Gzip compression of binary data against the \texttt{gzip} stdlib reference. \\

\texttt{lu\_\allowbreak{}factorization}
& max speedup
& Speed up dense LU factorization with partial pivoting against a LAPACK reference. \\

\end{longtable}
}%

\bigskip

{\setlength{\tabcolsep}{4pt}%
\renewcommand{\arraystretch}{1.15}%
\footnotesize
\begin{longtable}{@{}p{4.0cm} p{2.1cm} p{8.0cm}@{}}
\caption{%
Per-task descriptions for the \textbf{Heuristic Optimization} task group (35 tasks) evaluated under \datasetName{}.%
}
\label{tab:task-desc-alebench} \\
\toprule
\textbf{Task Name} & \textbf{Objective} & \textbf{Description} \\
\midrule
\endfirsthead

\multicolumn{3}{@{}l}{\textit{Table~\ref{tab:task-desc-alebench} (continued)}} \\
\toprule
\textbf{Task Name} & \textbf{Objective} & \textbf{Description} \\
\midrule
\endhead

\midrule
\multicolumn{3}{r@{}}{\textit{Continued on next page}} \\
\endfoot

\bottomrule
\endlastfoot

\texttt{ahc001} (Ad Placement)
& max score
& Place axis-parallel rectangular ads on a $10000{\times}10000$ canvas, each containing a fixed query point and matching a target area; maximize the sum of advertiser-satisfaction scores. \\

\texttt{ahc002} (Walking on Tiles)
& max score
& From start cell on a $50{\times}50$ tiled grid, walk a self-avoiding path (no tile reused); maximize the sum of cell scores along the path. \\

\texttt{ahc003} (Shortest Path Queries)
& max score
& On a $30{\times}30$ grid graph with unknown edge lengths, answer 1000 online shortest-path queries while learning edge lengths from noisy feedback. \\

\texttt{ahc004} (Cyclic Substring Cover)
& max score
& Pack given strings as horizontal/vertical cyclic substrings of an $N{\times}N$ matrix; maximize the number of strings covered while minimizing matrix usage. \\

\texttt{ahc005} (Patrol Route)
& min time
& On an $N{\times}N$ obstacle map with weighted road squares, design a closed walk from the start so that every road square is line-of-sight visible; minimize total travel time. \\

\texttt{ahc006} (Food Delivery)
& min length
& Out of 1000 delivery orders on an $801{\times}801$ grid, select 50 and produce a closed tour from the office that visits each pickup before its drop-off; minimize tour length. \\

\texttt{ahc007} (Online MST)
& min cost
& Online Steiner-MST: edges of an embedded graph arrive one by one with revealed length $\in[d_i,3d_i]$; decide on the fly whether to include each edge to ultimately span all terminals at minimum cost. \\

\texttt{ahc008} (Pet Herding)
& max satisfaction
& Control $M$ humans on a $30{\times}30$ grid for 300 turns, placing fences to isolate $N$ pets into pet-free regions; maximize the per-human area-isolation reward. \\

\texttt{ahc009} (Maze Walk)
& max P(reach)
& Output a single fixed action sequence $\in\{U,D,L,R\}^*$ that, under random execution slippage, maximizes the probability of reaching the goal in a $20{\times}20$ walled maze. \\

\texttt{ahc010} (Loop Tiles)
& max length
& Rotate $30{\times}30$ railroad-pattern tiles to form one closed loop of maximum total length. \\

\texttt{ahc011} (Sliding Tree Puzzle)
& max connectivity
& Slide tiles on an $N{\times}N$ board so the line patterns form one large connected tree; maximize tree size while staying within the move budget. \\

\texttt{ahc012} (Strawberry Cake)
& max satisfaction
& Cut a circular cake of radius $10^4$ with at most $K$ straight lines so that the strawberry counts in resulting pieces match a target multiset. \\

\texttt{ahc014} (RectJoin)
& max score
& In the RectJoin grid game, place dots and draw axis-aligned/diagonal rectangles repeatedly; maximize the weighted score of placed dots. \\

\texttt{ahc015} (Halloween Candy)
& max clusters
& In a $10{\times}10$ box where 100 candies of three flavors fall in a known order to random empty cells, choose a tilt direction before each candy to maximize same-flavor connected-component sizes. \\

\texttt{ahc016} (Graph Encoding)
& min error
& Given target $M$ and noise rate $\varepsilon$, design $M$ reference graphs $G_0,\dots,G_{M-1}$ that remain pairwise distinguishable after random edge flipping and label permutation. \\

\texttt{ahc017} (Edge Repair Schedule)
& min disruption
& Schedule the one-time repair of each edge of a planar weighted graph across $D$ days (at most $K$ repairs/day); minimize the total inflation in shortest-path distances during repairs. \\

\texttt{ahc019} (Polycube Silhouette)
& max match
& Combine polycube blocks into a single 3D object whose front and right silhouettes match two given 2D monochrome targets. \\

\texttt{ahc020} (Antenna Placement)
& max coverage
& Choose a connected subgraph and per-vertex broadcast power on a planar graph with weighted edges so that all $K$ residents are covered; minimize total power+edge cost. \\

\texttt{ahc021} (Pyramid Sort)
& min swaps
& Reorder balls in an $N$-tier pyramid via adjacent swaps so that the value at each position is at least its parent; minimize swap count. \\

\texttt{ahc024} (Map Compression)
& min size
& Re-tile an $n{\times}n$ multi-ward city map onto a smaller grid while preserving every ward's connectivity and inter-ward adjacency; minimize the resulting grid size. \\

\texttt{ahc025} (Weight Balancing)
& min imbalance
& Using only balance-scale comparisons of item subsets, partition $N$ items of unknown weight into $D$ groups of equal total weight within $Q$ queries. \\

\texttt{ahc026} (Cardboard Boxes)
& min cost
& Carry out $n$ uniquely-numbered boxes from $m$ stacks in ascending order using up to 5000 stack-relocation moves; minimize total move cost. \\

\texttt{ahc027} (Cleaning Robot)
& min dirt
& Design a closed cleaning route on an $N{\times}N$ walled grid; minimize the steady-state average dirtiness across cells with cell-specific dirt-accumulation rates. \\

\texttt{ahc028} (Keyboard Typing)
& min strokes
& Given an $N{\times}N$ keyboard with letters in cells and $M$ target words to type, output a finger trajectory whose visited letters contain every target word as a contiguous substring; minimize total moves. \\

\texttt{ahc030} (Polyomino Detection)
& min queries
& On an $N{\times}N$ island hiding $M$ polyomino oil-fields of known shape, locate every field via interactive cell/region drilling queries with noisy feedback. \\

\texttt{ahc031} (Event Hall Reservation)
& min penalty
& Partition a $W{\times}W$ hall into $N$ axis-aligned rectangles per day for $D$ days, each rectangle satisfying a daily area request; minimize unmet-area + cross-day partition-change penalties. \\

\texttt{ahc032} (Stamp Pressing)
& max sum
& On an $N{\times}N$ integer board, repeatedly press $3{\times}3$ stamps from $M$ stamp templates (mod $10^9{+}7$); maximize the final cell-sum. \\

\texttt{ahc034} (Dump Truck Leveling)
& min cost
& Drive a dump truck on an $N{\times}N$ height-grid (total height zero), loading and unloading dirt to flatten the field; minimize total transport cost. \\

\texttt{ahc035} (Seed Crossing)
& max yield
& Over a sequence of crossing rounds, schedule pairings of $2N(N{-}1)$ multi-trait seeds to evolve a high-quality final population. \\

\texttt{ahc039} (Purse-Seine Fishing)
& max catch
& Construct a rectilinear polygon in the plane that encloses as many mackerels and excludes as many sardines as possible, subject to a perimeter budget. \\

\texttt{ahc041} (Rooted Tree Forest)
& max attractiveness
& Partition a planar graph into rooted trees so that the weighted sum $\sum_v(h_v{+}1)A_v$ over depths and beauty values is maximized. \\

\texttt{ahc042} (Oni-Fukunokami Push)
& max remaining
& On an $N{\times}N$ board with Oni and Fukunokami tokens, repeatedly push entire rows/columns to drive Oni off the board while keeping as many Fukunokami as possible. \\

\texttt{ahc044} (Cleaning Schedule Automaton)
& min error
& Choose a per-employee finite-state-automaton transition $(a_i,b_i)$ so that, after a long simulated week sequence, employee $i$'s cleaning count converges to a target $T_i$. \\

\texttt{ahc045} (Group Exploration with Uncertain Coordinates)
& min length
& Cluster $N$ cities into tour groups of fixed sizes when each city's coordinates are only known up to a rectangular range, using a limited number of true-distance queries. \\

\texttt{ahc046} (Skating with Blocks)
& min actions
& On an $N{\times}N$ skating rink, visit target squares in order using \texttt{Move}, \texttt{Slide}, and block-placement actions; minimize total action count. \\

\end{longtable}
}%

\bigskip

{\setlength{\tabcolsep}{4pt}%
\renewcommand{\arraystretch}{1.15}%
\footnotesize
\begin{longtable}{@{}p{4.0cm} p{2.1cm} p{8.0cm}@{}}
\caption{%
Per-task descriptions for the \textbf{GPU Kernel Optimization} task group (4 tasks) evaluated under \datasetName{}.%
}
\label{tab:task-desc-gpu} \\
\toprule
\textbf{Task Name} & \textbf{Objective} & \textbf{Description} \\
\midrule
\endfirsthead

\multicolumn{3}{@{}l}{\textit{Table~\ref{tab:task-desc-gpu} (continued)}} \\
\toprule
\textbf{Task Name} & \textbf{Objective} & \textbf{Description} \\
\midrule
\endhead

\midrule
\multicolumn{3}{r@{}}{\textit{Continued on next page}} \\
\endfoot

\bottomrule
\endlastfoot

\texttt{grayscale}
& max throughput
& Triton/PyTorch kernel for RGB$\to$grayscale image conversion on H100/H200 GPUs. \\

\texttt{mla\_\allowbreak{}decode}
& max throughput
& Triton kernel for DeepSeek-V2/V3 multi-head latent-attention decoding. \\

\texttt{trimul}
& max throughput
& Triton kernel for the AlphaFold-3 triangle multiplicative update. \\

\texttt{vecadd}
& max throughput
& Triton/PyTorch kernel for FP16 vector addition on H100/H200 GPUs. \\

\end{longtable}
}%

\bigskip

\bigskip

{\setlength{\tabcolsep}{4pt}%
\renewcommand{\arraystretch}{1.15}%
\footnotesize
\begin{longtable}{@{}p{4.0cm} p{2.1cm} p{8.0cm}@{}}
\caption{%
Per-task descriptions for the \textbf{Constructive Search} task group (2 tasks) evaluated under \datasetName{}.%
}
\label{tab:task-desc-optimization} \\
\toprule
\textbf{Task Name} & \textbf{Objective} & \textbf{Description} \\
\midrule
\endfirsthead

\multicolumn{3}{@{}l}{\textit{Table~\ref{tab:task-desc-optimization} (continued)}} \\
\toprule
\textbf{Task Name} & \textbf{Objective} & \textbf{Description} \\
\midrule
\endhead

\midrule
\multicolumn{3}{r@{}}{\textit{Continued on next page}} \\
\endfoot

\bottomrule
\endlastfoot

\texttt{function\_\allowbreak{}minimization}
& $\min f$
& Black-box minimization of $f(x,y)=\sin(x)\cos(y)+\sin(xy)+(x^2{+}y^2)/20$ on $[-5,5]^2$; evolve a random-search seed into an adaptive optimizer. \\

\texttt{k\_\allowbreak{}module\_\allowbreak{}problem}
& max correct
& Discover the correct 4-component pipeline configuration ($5^4{=}625$ possibilities) when the only feedback per query is the count of correctly placed modules (deceptive landscape). \\

\end{longtable}
}%

\bigskip

{\setlength{\tabcolsep}{4pt}%
\renewcommand{\arraystretch}{1.15}%
\footnotesize
\begin{longtable}{@{}p{4.0cm} p{2.1cm} p{8.0cm}@{}}
\caption{%
Per-task descriptions for the \textbf{SR -- Bio Pop Growth} task group (24 tasks) evaluated under \datasetName{}.%
}
\label{tab:task-desc-sr-bio} \\
\toprule
\textbf{Task Name} & \textbf{Objective} & \textbf{Description} \\
\midrule
\endfirsthead

\multicolumn{3}{@{}l}{\textit{Table~\ref{tab:task-desc-sr-bio} (continued)}} \\
\toprule
\textbf{Task Name} & \textbf{Objective} & \textbf{Description} \\
\midrule
\endhead

\midrule
\multicolumn{3}{r@{}}{\textit{Continued on next page}} \\
\endfoot

\bottomrule
\endlastfoot

BPG0  & min fit error & Ground-truth: $\displaystyle 0.954\,(1-P/96.9)\,P + 0.954\,P^{0.333}$. \\

BPG1  & min fit error & Ground-truth: $\displaystyle 0.316\,P\,e^{-0.0541 t} + 0.316\,P^{2}/(9.87\,P+1)$. \\

BPG2  & min fit error & Ground-truth: $\displaystyle 0.257\,P\sin(0.722\,t) + 0.115\,P\,e^{-0.0304 t}$. \\

BPG3  & min fit error & Ground-truth: $\displaystyle 0.845\,(-1+P/5.12)(1-P/34.4)\,P + 0.845\,(1-e^{-0.0969 P})\,P$. \\

BPG4  & min fit error & Ground-truth: $\displaystyle 0.173\,(1-P/48.5)\,P + 0.173\,P/(1+e^{-1.52(-0.924+P)})$. \\

BPG5  & min fit error & Ground-truth: $\displaystyle 0.92\,(1-P/84)\,P + 0.92\,P^{2}/(7.53\,P+1)$. \\

BPG6  & min fit error & Ground-truth: $\displaystyle -6.22\cdot 1.41\,P + 0.858\,(1-P/79.2)\,P + 0.858\,P^{0.333} + 0.858\,P$. \\

BPG7  & min fit error & Ground-truth: $\displaystyle 0.721\,(-1+P/6.65)(1-P/13.5)\,P + 0.721\,(1-P/13.5)\,P + 0.721\,P^{0.333}$. \\

BPG8  & min fit error & Ground-truth: $\displaystyle 0.991\,(1-P/39)\,P + 0.991\,P^{0.333} + 0.991\,P$. \\

BPG9  & min fit error & Ground-truth: $\displaystyle 0.17\,(-1+P/1.05)(1-P/10.2)\,P + 0.17\,(1-P/10.2)\,P + 0.17\,(1-e^{-0.0971 P})\,P$. \\

BPG10 & min fit error & Ground-truth: $\displaystyle 0.101\,P^{0.333} + 0.101\,P$. \\

BPG11 & min fit error & Ground-truth: $\displaystyle 0.712\,(1-P/68.9)\,P + 0.712\,P^{0.333} + 0.712\,P\,e^{-0.0346 t}$. \\

BPG12 & min fit error & Ground-truth: $\displaystyle 0.877\,P\sin(0.567\,t) + 0.701\,(1-P/65.8)\,P$. \\

BPG13 & min fit error & Ground-truth: $\displaystyle 0.201\,(-1+P/5.64)(1-P/10.2)\,P + 0.201\,P + 0.201\,P/(1+e^{-5.64(-0.634+P)})$. \\

BPG14 & min fit error & Ground-truth: $\displaystyle 0.114\,(1-P/40.7)\,P + 0.114\,(1-e^{-0.0837 P})\,P + 0.114\,P\,e^{-0.0837 t}$. \\

BPG15 & min fit error & Ground-truth: $\displaystyle 0.487\,P^{0.333} + 0.487\,P\,e^{-0.0858 t}$. \\

BPG16 & min fit error & Ground-truth: $\displaystyle 0.868\,(-1+P/6.06)(1-P/14.2)\,P + 0.868\,P^{0.333} + 0.868\,P$. \\

BPG17 & min fit error & Ground-truth: $\displaystyle 0.769\,(-1+P/8.67)(1-P/14.3)\,P + 0.769\,P^{0.333}$. \\

BPG18 & min fit error & Ground-truth: $\displaystyle 0.477\,P\sin(0.776\,t) + 0.445\,(1-P/51.1)\,P + 0.445\,P$. \\

BPG19 & min fit error & Ground-truth: $\displaystyle 0.421\,(1-P/84.4)\,P + 0.421\,P/t^{2.59}$. \\

BPG20 & min fit error & Ground-truth: $\displaystyle 0.139\,(-1+P/8.04)(1-P/70)\,P + 0.139\,(1-P/70)\,P + 0.139\,P/(1+e^{-8.04(-0.589+P)})$. \\

BPG21 & min fit error & Ground-truth: $\displaystyle 0.14\,(-1+P/4.53)(1-P/78.5)\,P + 0.14\,P/t^{4.53}$. \\

BPG22 & min fit error & Ground-truth: $\displaystyle 0.118\,(1-e^{-0.0272 P})\,P + 0.118\,P\,e^{-0.0272 t}$. \\

BPG23 & min fit error & Ground-truth: $\displaystyle 0.598\,(1-P/32.9)\,P + 0.598\,(1-e^{-0.0768 P})\,P$. \\

\end{longtable}
}%

\bigskip

{\setlength{\tabcolsep}{4pt}%
\renewcommand{\arraystretch}{1.15}%
\footnotesize
\begin{longtable}{@{}p{4.0cm} p{2.1cm} p{8.0cm}@{}}
\caption{%
Per-task descriptions for the \textbf{SR -- Chem Reaction} task group (12 tasks) evaluated under \datasetName{}.%
}
\label{tab:task-desc-sr-chem} \\
\toprule
\textbf{Task Name} & \textbf{Objective} & \textbf{Description} \\
\midrule
\endfirsthead

\multicolumn{3}{@{}l}{\textit{Table~\ref{tab:task-desc-sr-chem} (continued)}} \\
\toprule
\textbf{Task Name} & \textbf{Objective} & \textbf{Description} \\
\midrule
\endhead

\midrule
\multicolumn{3}{r@{}}{\textit{Continued on next page}} \\
\endfoot

\bottomrule
\endlastfoot

CRK0  & min fit error & Ground-truth: $\displaystyle -0.19\,A^{2} + 0.19\,A^{2}/(0.75\,A^{4}+1)$. \\

CRK1  & min fit error & Ground-truth: $\displaystyle -0.773\,A^{2} - 0.773\,A + 0.773\,\cos(\log(A+1))$. \\

CRK2  & min fit error & Ground-truth: $\displaystyle -0.195\,A + 0.195\,\cos(\log(A+1))$. \\

CRK3  & min fit error & Ground-truth: $\displaystyle -0.548\,A^{2} - 0.548\,A\,e^{-0.548 t} + 0.548\,\cos(\log(A+1))$. \\

CRK4  & min fit error & Ground-truth: $\displaystyle -0.115\,A^{2} + 0.115\,A\,\log(0.257\,t+1)$. \\

CRK5  & min fit error & Ground-truth: $\displaystyle -0.325\,\sqrt{A} + 0.325\,A^{0.333}$. \\

CRK6  & min fit error & Ground-truth: $\displaystyle -0.447\,A\,e^{-0.447 t} + 0.447\,\sin(\sqrt{A})$. \\

CRK7  & min fit error & Ground-truth: $\displaystyle -0.732\,A\,e^{-0.732 t} + 0.732\,\cos(\log(A+1))$. \\

CRK8  & min fit error & Ground-truth: $\displaystyle -0.679\,A^{2} - 0.679\,A + 0.679\,\sin(\log(A+1))$. \\

CRK9  & min fit error & Ground-truth: $\displaystyle -0.26\,\sqrt{A} + 0.26\,\cos(\log(A+1))$. \\

CRK10 & min fit error & Ground-truth: $\displaystyle -0.175\,A^{2} + 0.175\,\sin(\log(A+1))$. \\

CRK11 & min fit error & Ground-truth: $\displaystyle -0.882\,A^{2} + 0.882\,\sin(\sqrt{A})$. \\

\end{longtable}
}%

\bigskip

{\setlength{\tabcolsep}{4pt}%
\renewcommand{\arraystretch}{1.15}%
\footnotesize
\begin{longtable}{@{}p{4.0cm} p{2.1cm} p{8.0cm}@{}}
\caption{%
Per-task descriptions for the \textbf{SR -- Physics Oscillation} task group (44 tasks) evaluated under \datasetName{}.%
}
\label{tab:task-desc-sr-phys} \\
\toprule
\textbf{Task Name} & \textbf{Objective} & \textbf{Description} \\
\midrule
\endfirsthead

\multicolumn{3}{@{}l}{\textit{Table~\ref{tab:task-desc-sr-phys} (continued)}} \\
\toprule
\textbf{Task Name} & \textbf{Objective} & \textbf{Description} \\
\midrule
\endhead

\midrule
\multicolumn{3}{r@{}}{\textit{Continued on next page}} \\
\endfoot

\bottomrule
\endlastfoot

PO0  & min fit error & Ground-truth: $\displaystyle F_0\sin t - \beta\sin v - \omega_0^{2}x^{3} - \omega_0^{2}x\,e^{-|x|}$. \\

PO1  & min fit error & Ground-truth: $\displaystyle F_0\sin t - \omega_0^{2}x - \omega_0^{2}x\,e^{-|x|}$. \\

PO2  & min fit error & Ground-truth: $\displaystyle -\alpha v^{3} - \mu(1-x^{2})v - \omega_0^{2}x - \omega_0^{2}x\,e^{-|x|}$. \\

PO3  & min fit error & Ground-truth: $\displaystyle F_0\sin t - \beta\sin v - 2\beta v$. \\

PO4  & min fit error & Ground-truth: $\displaystyle F_0\sin t - \alpha v^{3} - \omega_0^{2}(\gamma|v|^{0.333}+1)x - \omega_0^{2}x$. \\

PO5  & min fit error & Ground-truth: $\displaystyle -\beta\sin v - 2\beta v - \omega_0^{2}(\gamma|v|^{0.333}+1)x - \omega_0^{2}x^{3} - \omega_0^{2}x$. \\

PO6  & min fit error & Ground-truth: $\displaystyle -\beta\log(|v|+1) - 2\beta v - \omega_0^{2}x^{3}$. \\

PO7  & min fit error & Ground-truth: $\displaystyle -\alpha v^{3} - \beta|v|^{0.333} - \omega_0^{2}x^{3}$. \\

PO8  & min fit error & Ground-truth: $\displaystyle -\beta|v|^{0.333} - \omega_0^{2}x^{3}$. \\

PO9  & min fit error & Ground-truth: $\displaystyle F_0\sin t - \mu(1-x^{2})v - \omega_0^{2}(\gamma|v|^{0.333}+1)x - \omega_0^{2}x$. \\

PO10 & min fit error & Ground-truth: $\displaystyle F_0\sin t - \omega_0^{2}(\gamma t+1)x - \omega_0^{2}x^{3} - \omega_0^{2}x$. \\

PO11 & min fit error & Ground-truth: $\displaystyle -\beta\sin v - \omega_0^{2}(\gamma t+1)x - \omega_0^{2}x^{3}$. \\

PO12 & min fit error & Ground-truth: $\displaystyle F_0\sin t - \alpha v^{3} - \beta|v|^{0.333} - \omega_0^{2}(\gamma t+1)x - \omega_0^{2}x$. \\

PO13 & min fit error & Ground-truth: $\displaystyle F_0\sin t - \mu(1-x^{2})v - \omega_0^{2}(\gamma|v|^{0.333}+1)x$. \\

PO14 & min fit error & Ground-truth: $\displaystyle F_0\sin t - \beta\log(|v|+1) - \beta\sin v - 2\beta v - \mu(1-x^{2})v$. \\

PO15 & min fit error & Ground-truth: $\displaystyle F_0\sin t - \omega_0^{2}(\gamma|v|^{0.333}+1)x - \omega_0^{2}x - \omega_0^{2}x\,e^{-|x|}$. \\

PO16 & min fit error & Ground-truth: $\displaystyle F_0\sin t - \beta\sin(x)\,v - \beta\sin v - \omega_0^{2}x^{3}$. \\

PO17 & min fit error & Ground-truth: $\displaystyle F_0\sin t - \beta\sin(x)\,v - 2\beta v - \omega_0^{2}x$. \\

PO18 & min fit error & Ground-truth: $\displaystyle -\beta\sin(x)\,v - \omega_0^{2}x$. \\

PO19 & min fit error & Ground-truth: $\displaystyle -2\beta v - \omega_0^{2}x\,e^{-|x|}$. \\

PO20 & min fit error & Ground-truth: $\displaystyle -\alpha v^{3} - \beta\log(|v|+1) - 2\beta v - \mu(1-v^{2})v - \omega_0^{2}(\gamma|v|^{0.333}+1)x$. \\

PO21 & min fit error & Ground-truth: $\displaystyle F_0\sin t - \beta\sin(x)\,v$. \\

PO22 & min fit error & Ground-truth: $\displaystyle -2\beta v - \beta e^{-|x|} v - \mu(1-x^{2})v - \omega_0^{2}x^{3}$. \\

PO23 & min fit error & Ground-truth: $\displaystyle F_0\sin t - \beta\log(|v|+1) - \omega_0^{2}x\,e^{-|x|}$. \\

PO24 & min fit error & Ground-truth: $\displaystyle F_0\sin t - \alpha v^{3} - \beta\log(|v|+1)$. \\

PO25 & min fit error & Ground-truth: $\displaystyle F_0\sin t - \beta\sin v$. \\

PO26 & min fit error & Ground-truth: $\displaystyle F_0\sin t - \beta\log(|v|+1) - 2\beta v - \omega_0^{2}x^{3}$. \\

PO27 & min fit error & Ground-truth: $\displaystyle F_0\sin t - \alpha v^{3} - 2\beta v - \beta e^{-|v|} v$. \\

PO28 & min fit error & Ground-truth: $\displaystyle -2\beta v - \omega_0^{2}(\gamma|v|^{0.333}+1)x - \omega_0^{2}x^{3} - \omega_0^{2}x$. \\

PO29 & min fit error & Ground-truth: $\displaystyle -\mu(1-x^{2})v - \omega_0^{2}(\gamma t+1)x - \omega_0^{2}x^{3}$. \\

PO30 & min fit error & Ground-truth: $\displaystyle -\alpha v^{3} - \beta\sin(x)\,v - \beta\sin v - \omega_0^{2}x^{3}$. \\

PO31 & min fit error & Ground-truth: $\displaystyle -\omega_0^{2}(\gamma|v|^{0.333}+1)x - \omega_0^{2}x^{3}$. \\

PO32 & min fit error & Ground-truth: $\displaystyle F_0\sin t - \alpha v^{3} - \beta e^{-|v|} v - \omega_0^{2}x^{3}$. \\

PO33 & min fit error & Ground-truth: $\displaystyle -2\beta v - \mu(1-v^{2})v - \omega_0^{2}(\gamma t+1)x - \omega_0^{2}x$. \\

PO34 & min fit error & Ground-truth: $\displaystyle -2\beta v - \mu(1-v^{2})v - \omega_0^{2}(\gamma|v|^{0.333}+1)x$. \\

PO35 & min fit error & Ground-truth: $\displaystyle F_0\sin t - \beta\sin v - \omega_0^{2}(\gamma|v|^{0.333}+1)x$. \\

PO36 & min fit error & Ground-truth: $\displaystyle F_0\sin t - \beta e^{-|x|} v$. \\

PO37 & min fit error & Ground-truth: $\displaystyle F_0\sin t - \alpha v^{3} - 2\beta v - \omega_0^{2}(\gamma t+1)x$. \\

PO38 & min fit error & Ground-truth: $\displaystyle -\beta\sin v - \mu(1-x^{2})v - \omega_0^{2}x\,e^{-|x|}$. \\

PO39 & min fit error & Ground-truth: $\displaystyle F_0\sin t - \alpha v^{3} - \beta e^{-|x|} v - \mu(1-v^{2})v$. \\

PO40 & min fit error & Ground-truth: $\displaystyle F_0\sin t - \beta|v|^{0.333} - \omega_0^{2}(\gamma|v|^{0.333}+1)x - \omega_0^{2}x^{3} - \omega_0^{2}x$. \\

PO41 & min fit error & Ground-truth: $\displaystyle -\mu(1-x^{2})v - \omega_0^{2}x\,e^{-|x|}$. \\

PO42 & min fit error & Ground-truth: $\displaystyle F_0\sin t - \alpha v^{3} - \beta\sin(x)\,v - 2\beta v$. \\

PO43 & min fit error & Ground-truth: $\displaystyle F_0\sin t - \beta\sin(x)\,v - 2\beta v - \mu(1-x^{2})v - \omega_0^{2}x\,e^{-|x|}$. \\

\end{longtable}
}%

\bigskip

{\setlength{\tabcolsep}{4pt}%
\renewcommand{\arraystretch}{1.15}%
\footnotesize
\begin{longtable}{@{}p{5.6cm} p{2.1cm} p{6.6cm}@{}}
\caption{%
Per-task descriptions for the \textbf{Competitive Programming (Frontier-CS)} task group (172 tasks) evaluated under \datasetName{}.%
}
\label{tab:task-desc-frontier-cs} \\
\toprule
\textbf{Task Name} & \textbf{Objective} & \textbf{Description} \\
\midrule
\endfirsthead

\multicolumn{3}{@{}l}{\textit{Table~\ref{tab:task-desc-frontier-cs} (continued)}} \\
\toprule
\textbf{Task Name} & \textbf{Objective} & \textbf{Description} \\
\midrule
\endhead

\midrule
\multicolumn{3}{r@{}}{\textit{Continued on next page}} \\
\endfoot

\bottomrule
\endlastfoot

FCS-0: Pack the Polyominoes (Reflections Allowed) & $\min$ area & Minimize the bounding-rectangle area for a polyomino-packing instance allowing rotations and reflections. \\

FCS-1: Treasure Packing & judge score & Treasure-packing constructive optimization. \\

FCS-2: Permutation & judge score & Construct a permutation satisfying problem-specific constraints. \\

FCS-3: Ring Lamp Permutation Recovery (interactive) & $\min$ queries & Recover a hidden cyclic permutation via lamp-toggle queries with adjacency feedback. \\

FCS-4: Matrix $k$-th Smallest (interactive) & $\min$ queries & Identify the $k$-th smallest element of a hidden matrix through interactive queries. \\

FCS-5: Hamiltonian Path Challenge & judge score & Construct or approximate a Hamiltonian path on a given graph instance. \\

FCS-6: World Map & judge score & Constructive world-map design. \\

FCS-7: Build a Computer & judge score & Combinatorial construction problem (``Build a Computer''). \\

FCS-8: The Empress & judge score & ``The Empress'' constructive optimization problem. \\

FCS-9: Tree Permutation Sort via Matchings & $\min$ swaps & Sort a tree-labeled permutation to identity using minimum-matching swaps. \\

FCS-10: Tree Distance & judge score & Tree-distance computation problem. \\

FCS-11: Palindrome Path & judge score & Find a palindromic path in a graph or grid. \\

FCS-13: Robot Explosion on Grid (interactive) & $\le 3000$ steps & Detonate a grid robot in $\le 3000$ steps using interactive cell-marking queries. \\

FCS-14: Hidden Cycle Length (interactive) & $\min$ queries & Determine the length of a hidden cycle through interactive queries. \\

FCS-15: Permutation Minimization via 3-Part Cut/Swap & $\min$ lex order & Minimize lexicographic permutation order using $\le 4n$ cut-and-swap operations. \\

FCS-16: Identify Chord & judge score & Identify a hidden chord on a circle through queries. \\

FCS-17: Permutation (Interactive) & judge score & Interactive permutation-recovery variant. \\

FCS-22: A+B Problem (Hard) & judge score & Hard variant of the A+B problem (high-precision / large-input). \\

FCS-23: A = B (Sequence Transformation) & judge score & Transform sequence $A$ into $B$ via constrained operations. \\

FCS-24: Almost Monochromatic Permutation & judge score & Construct an ``almost monochromatic'' permutation. \\

FCS-25: Graph Connectivity via Neighborhood Queries (interactive) & $\min$ queries & Decide graph connectivity using only neighborhood queries. \\

FCS-26: OgreSort & judge score & Custom comparison-sort variant (``OgreSort''). \\

FCS-27: No-Axis-Rectangle Point Placement & $\max$ points & Place the maximum number of points in an $n{\times}m$ grid with no axis-parallel rectangle of any 4 of them. \\

FCS-28: Hacking the Project & judge score & ``Hacking the Project'' constructive problem. \\

FCS-30: Hidden Mole in Tree (interactive) & $\min$ queries & Locate a hidden mole on a tree via interactive vertex queries. \\

FCS-33: Permutation (Modified Version) & judge score & Modified-version of an interactive permutation problem. \\

FCS-35: Unique Element in Doubled Array (interactive) & $\min$ queries & Identify the unique element in a doubled array via comparison queries. \\

FCS-36: Hack! & judge score & ``Hack!'' constructive challenge. \\

FCS-40: Interactive RBS & judge score & Recover a hidden balanced-bracket sequence interactively. \\

FCS-41: Brilliant Umbrella Sequence & judge score & Construct a strictly increasing sequence whose consecutive GCDs are also strictly increasing. \\

FCS-42: Unit Square Packing with Rotations & $\min$ area & Pack $n$ unit squares allowing arbitrary rotation in the smallest enclosing square. \\

FCS-43: Bigger Sokoban 40k & judge score & Larger-scale Sokoban-style puzzle (40k cells). \\

FCS-44: Traveling Santa with Carrot Constraint & judge score & Tour problem with carrot-pickup constraints. \\

FCS-45: Balanced Graph Partitioning (DIMACS10) & $\min$ cut & Balanced partitioning of DIMACS10 graphs minimizing edge cut. \\

FCS-46: Job Shop Scheduling (JSPLIB) & $\min$ makespan & Classical job-shop scheduling on JSPLIB instances. \\

FCS-47: 2D Rectangular Knapsack with 90\textdegree{} Rotations & $\max$ value & 2D rectangle knapsack with 90$^\circ$ rotations allowed. \\

FCS-48: Sphere Packing in a Cube & $\min$ cube & Pack $n$ unit spheres inside the smallest enclosing cube. \\

FCS-50: Weighted Set Cover & $\min$ cost & Minimum-cost weighted set cover. \\

FCS-52: Geemu & judge score & Game-state constructive problem (``Geemu''). \\

FCS-53: G2. Inter Active (Hard Version) & judge score & Hard interactive variant ``G2. Inter Active''. \\

FCS-54: Centroid Guess & $\min$ queries & Interactive centroid-guessing on a tree. \\

FCS-57: Rooted Tree Identification via Path-Sum Queries (interactive) & $\min$ queries & Identify a rooted tree using path-sum queries. \\

FCS-58: Inverse Counting Path & judge score & Inverse-counting path-construction problem. \\

FCS-59: Limited Shuffle Restoring & judge score & Restore an array using a limited number of shuffles. \\

FCS-60: Probing the Disk (interactive) & $\min$ queries & Detect a hidden disk's center and radius via line-segment light probes. \\

FCS-61: Let's Go! New Adventure & judge score & ``Let's Go! New Adventure'' constructive problem. \\

FCS-62: Ball Moving Game & judge score & Constructive ball-moving puzzle. \\

FCS-63: Space Thief & judge score & ``Space Thief'' constructive problem. \\

FCS-64: Subset Sum Closest to Target & $\min$ gap & Subset-sum closest to a target value. \\

FCS-68: Pen Ink Selection (interactive) & $\min$ queries & Interactive pen-ink-selection problem. \\

FCS-69: Distinct Magic Words (Spell Power) & $\max$ distinct & Maximize the number of distinct substrings (``magic words'') in a string. \\

FCS-70: Treasure Hunt & judge score & ``Treasure Hunt'' constructive problem. \\

FCS-72: Rush-Hour Puzzle & $\max$ steps & On a $6{\times}6$ Rush-Hour puzzle, maximize the minimum steps required for any solution. \\

FCS-73: Inversion & judge score & ``Inversion'' constructive problem. \\

FCS-75: Black and White & judge score & ``Black and White'' constructive problem. \\

FCS-77: Izzy Wager Predictions & $\min$ wrong & Minimize wrong guesses across 10{,}000 interactive prediction rounds. \\

FCS-79: H. Hack & judge score & ``H. Hack'' constructive problem. \\

FCS-80: Indiana Jones and the Uniform Cave & judge score & ``Indiana Jones'' interactive cave-traversal problem. \\

FCS-81: Stone Slate String Recovery & judge score & Recover a hidden string written on a stone slate. \\

FCS-82: Permutation Recovery via OR Queries & $\min$ queries & Recover a hidden permutation in $\le 4269$ bitwise-OR queries. \\

FCS-83: Completely Multiplicative Function Partial-Sum Minimization & $\min$ partial sum & For a completely multiplicative $f(\cdot)$, minimize $\max_k|\sum_{i\le k} f(i)|$. \\

FCS-85: Maze & judge score & ``Maze'' constructive problem. \\

FCS-86: Hidden Tree & judge score & Recover a hidden tree via queries. \\

FCS-87: Graph Coloring & judge score & Graph-coloring decision/optimization. \\

FCS-89: Tree Reconstruction via Steiner-Membership Queries & $\min$ queries & Reconstruct a tree using Steiner-set membership queries. \\

FCS-93: Greedy & judge score & ``Greedy'' constructive problem. \\

FCS-101: Circuit & judge score & ``Circuit'' constructive problem. \\

FCS-104: Dishonest Student Attendance (interactive) & $\min$ queries & Interactive ``dishonest-student attendance'' problem. \\

FCS-106: Hidden Bipartite Graph & judge score & Recover a hidden bipartite graph. \\

FCS-107: F. Guess Divisors Count & $\min$ queries & Guess a number's divisor count using interactive queries. \\

FCS-108: Ring Lock with Arc Alignment (interactive) & $\min$ ops & Open a ring lock by aligning arcs with the fewest interactive operations. \\

FCS-109: Knight's Longest Tour & $\max$ length & Find the longest knight's tour on a chess sub-board. \\

FCS-110: 8x14 Digit Grid (Maximize Readable Numbers) & $\max$ count & Fill an $8{\times}14$ digit grid to maximize the number of 8-directional readable numerals. \\

FCS-111: Distinct Pairwise XOR Set & $\max$ size & Construct a set whose pairwise XORs are all distinct. \\

FCS-112: SphereSpread & judge score & ``SphereSpread'' constructive problem. \\

FCS-113: Three Baskets Center-Ball Moves & judge score & Constructive ball-distribution between three baskets. \\

FCS-117: Line & judge score & ``Line'' constructive problem. \\

FCS-119: Operators & judge score & Constructive operator-insertion puzzle. \\

FCS-120: Da Bai (interactive) & $\min$ queries & ``Da Bai'' interactive problem. \\

FCS-121: DNA Matching & judge score & DNA-matching constructive problem. \\

FCS-122: Hidden Editor Line-Width Discovery (interactive) & $\min$ queries & Discover a hidden editor's line width via interactive queries. \\

FCS-123: Hidden Integer via Set-Membership Queries & $\min$ queries & Recover a hidden integer through set-membership queries. \\

FCS-124: AveragePermutation & judge score & ``AveragePermutation'' constructive problem. \\

FCS-125: Hidden Mineral Pair Discovery (Device Query) & $\min$ queries & Discover a hidden mineral pair via device queries. \\

FCS-127: Hidden Diamond Box (interactive) & $\min$ queries & Locate a hidden diamond inside a box through interactive probes. \\

FCS-132: Robot-based Chairman Location Finder (interactive) & $\min$ queries & Locate a hidden chairman on a graph using a movable robot. \\

FCS-133: Line Segment Brush Area & $\max$ area & Maximize the brush-stroke area covered by a set of line segments. \\

FCS-134: Guess Number & $\min$ queries & Interactive number-guessing variant. \\

FCS-135: Interactive Problem (Standard Solutions Won't Work) & judge score & Adversarial interactive problem requiring non-standard strategy. \\

FCS-137: Kangaroos & judge score & ``Kangaroos'' constructive problem. \\

FCS-138: Fabulous Fungus Frenzy & judge score & ``Fabulous Fungus Frenzy'' constructive problem. \\

FCS-140: Mineral Deposits (interactive) & $\min$ queries & Interactive mineral-deposit identification problem. \\

FCS-141: Bakery Survey & judge score & ``Bakery Survey'' constructive problem. \\

FCS-142: Ball Game & judge score & ``Ball Game'' constructive problem. \\

FCS-143: Texas Hold'em Training (interactive) & judge score & Interactive Texas-Hold'em training problem. \\

FCS-144: Find Median & $\min$ queries & Interactively find the median of a hidden array. \\

FCS-145: Meituan Cup: Number Loop & judge score & Meituan Cup ``Number Loop'' constructive problem. \\

FCS-147: Ad Rectangle Placement & max score & $10000{\times}10000$ ad-rectangle placement (AHC001-style). \\

FCS-148: Tile Path Score Maximization & max score & $50{\times}50$ tile-path score maximization (AHC002-style). \\

FCS-149: Grid Shortest Path with Unknown Edges & max score & Online shortest-path queries on a grid with unknown edge weights. \\

FCS-150: Torus Matrix Recovery from Substrings & judge score & Recover a torus-matrix matching given cyclic substrings. \\

FCS-151: Patrol Route Minimization & $\min$ time & Patrol-route minimization on weighted road grid (AHC005-style). \\

FCS-152: Food Delivery Route with 50 Orders & $\min$ length & Select 50 of 1000 orders on a cyclic delivery route (AHC006-style). \\

FCS-153: Online MST with Edge Length Uncertainty & $\min$ cost & Online MST under edge-length uncertainty (AHC003-style). \\

FCS-154: Pet Herding via Partitions & max score & Pet herding on a $30{\times}30$ grid via partitions (AHC008-style). \\

FCS-155: Robust Commute String & max score & Robust action-sequence design under stochastic execution (AHC004-style). \\

FCS-156: Loop Line Toy Train & max score & Toy-train loop-line track-laying (AHC009-style). \\

FCS-157: Sliding Puzzle Tree Target & max score & Sliding-tile puzzle to form a target spanning tree (AHC011-style). \\

FCS-158: Cake Cutting with Strawberries & judge score & Cut a circular cake with $\le K$ lines to match a strawberry-count target (AHC012-style). \\

FCS-159: RectJoin Grid Puzzle & max score & RectJoin grid-puzzle constructive problem (AHC014-style). \\

FCS-160: Halloween Candy Tilt (10x10) & max clusters & Tilt-controlled candy-clustering on a $10{\times}10$ box (AHC015-style). \\

FCS-161: TV Network (Planar Graph) & judge score & Planar-graph TV-network construction (AHC010-style). \\

FCS-162: Pyramid Ball Swap & $\min$ swaps & Pyramid-ball reorder via adjacent swaps (AHC021-style). \\

FCS-163: Map Tile Coloring Minimization & $\min$ size & Map-tile coloring/compression preserving adjacencies (AHC024-style). \\

FCS-164: Cardboard Box Carry-Out & $\min$ cost & Cardboard-box carry-out with minimum lifting energy (AHC026-style). \\

FCS-165: Lucky String Keyboard & judge score & Keyboard-layout design for lucky-string typing. \\

FCS-166: Dump Truck Ground Leveling & $\min$ cost & Dump-truck ground leveling on an $N{\times}N$ grid (AHC034-style). \\

FCS-167: Purse Seine Fishing & max score & Rectilinear-polygon mackerel/sardine separation (AHC039-style). \\

FCS-168: Rooted Christmas Tree Beauty & max score & Rooted-tree beauty maximization. \\

FCS-169: Oni wa Soto, Fuku wa Uchi (Setsubun board game) & max score & Setsubun-themed Oni/Fuku push-board game. \\

FCS-170: Cleaning Duty Assignment & judge score & Cleaning-duty assignment under per-employee targets. \\

FCS-171: Skating Rink Block Placement & $\min$ ops & Skating-rink block-placement order-of-visit problem (AHC046-style). \\

FCS-174: Graph 3-Coloring & judge score & 3-coloring of a given graph. \\

FCS-175: 3-SAT & judge score & 3-SAT instance (Frontier-CS variant). \\

FCS-176: 3-SAT & judge score & 3-SAT instance (different size class). \\

FCS-177: Graph 3-Coloring & judge score & 3-coloring instance (different size class). \\

FCS-178: 3-SAT & judge score & 3-SAT instance (third size class). \\

FCS-179: Subset Sum & judge score & Subset-sum instance. \\

FCS-180: Graph Isomorphism & judge score & Graph-isomorphism instance. \\

FCS-181: Binary Quadratic Assignment Problem & judge score & Binary-QAP instance. \\

FCS-182: Vertex Cover Challenge & judge score & Vertex-cover challenge. \\

FCS-183: Maximum Independent Set Challenge & judge score & Maximum-independent-set challenge. \\

FCS-184: Maximum Independent Set Challenge & judge score & Maximum-independent-set challenge (different instance). \\

FCS-185: Maximum Clique Challenge & judge score & Maximum-clique challenge. \\

FCS-186: Graph Coloring Challenge & judge score & Graph-coloring challenge. \\

FCS-187: Clique Cover Challenge & judge score & Clique-cover challenge. \\

FCS-188: LCS Challenge (Approximation) & judge score & Approximate longest-common-subsequence challenge. \\

FCS-189: Edit Distance Challenge (Approximation) & judge score & Approximate edit-distance challenge. \\

FCS-192: Max-Cut & judge score & Max-cut instance. \\

FCS-193: Max-2-SAT & judge score & Max-2-SAT instance. \\

FCS-203: Chameleon & judge score & ``Chameleon'' constructive problem. \\

FCS-205: Sequence Transformation & judge score & Constrained sequence-transformation problem. \\

FCS-207: Efficient Sorting & judge score & Cost-bounded sorting variant. \\

FCS-209: Hidden Weights & judge score & Recover hidden item weights via comparison queries. \\

FCS-210: Military Exercise: Fighter Scheduling \& Base Strikes & judge score & Joint fighter scheduling and base-strike planning. \\

FCS-211: Communication Robots & judge score & ``Communication Robots'' constructive problem. \\

FCS-212: I Wanna Cross the Grid & judge score & ``I Wanna Cross the Grid'' constructive problem. \\

FCS-213: Sequence Shift (Moqueve) & judge score & ``moqueve'' sequence-shift problem. \\

FCS-214: Sequence Reversal (Requese) & judge score & ``requese'' sequence-reversal problem. \\

FCS-217: Super Dango Maker & judge score & ``Super Dango Maker'' constructive problem. \\

FCS-220: Playing Around the Table & judge score & ``Playing Around the Table'' constructive problem. \\

FCS-222: Hedgehog Graph & judge score & ``Hedgehog'' graph-construction problem. \\

FCS-225: Permutation Set Merging & judge score & Permutation-set merging problem. \\

FCS-227: 4-Way Partition: max LIS+LDS+LIS+LDS & $\max$ sum & 4-way partition maximizing the combined LIS$+$LDS$+$LIS$+$LDS lengths. \\

FCS-228: 01-String: Substrings with $|0|=|1|^{2}$ & judge score & Count $\{0,1\}$-string substrings where $\#0=(\#1)^2$. \\

FCS-229: LIS after Interval Shifts (Cheated LIS) & $\max$ LIS & Maximize LIS after $\le 10$ interval shifts. \\

FCS-231: Differentiating Games & judge score & ``Differentiating Games'' constructive problem. \\

FCS-233: Snake & judge score & ``Snake'' constructive problem. \\

FCS-239: Shortest 3-Hop DAG Augmentation & $\min$ edges & Add minimum edges to a $0{\to}n$ chain so every $v{<}u$ has a $\le 3$-edge path. \\

FCS-241: Boolean Expression Construction (AND/OR from Truth Table) & judge score & Construct an AND/OR Boolean expression matching a given truth table. \\

FCS-243: Hidden Position Determination on Grid Map (interactive) & $\min$ queries & Locate a hidden position on a grid map via interactive queries. \\

FCS-245: Asesino & judge score & ``Asesino'' constructive problem. \\

FCS-247: Sequence Swap to Reach $B$ & judge score & Transform $A$ into $B$ via swap-and-adjust ops $(A_i,A_j)\to(A_j{-}1,A_i{+}1)$. \\

FCS-248: Drone Delivery & judge score & ``Drone Delivery'' constructive problem. \\

FCS-249: X-OR & judge score & ``X-OR'' constructive problem. \\

FCS-252: Hotel & judge score & ``Hotel'' constructive problem. \\

FCS-253: Roads & judge score & ``Roads'' constructive problem. \\

FCS-254: Pepe Racing & judge score & ``Pepe Racing'' constructive problem. \\

FCS-255: Magnets & judge score & ``Magnets'' constructive problem. \\

FCS-256: Palindromic Paths & judge score & ``Palindromic Paths'' constructive problem. \\

FCS-257: Omkar and Modes & judge score & ``Omkar and Modes'' constructive problem. \\

FCS-258: Network Synchronization: Finding Dual Anomalies & judge score & Network-synchronization dual-anomaly detection. \\

\end{longtable}
}%

\bigskip


\end{document}